%% file: main.tex
\pdfoutput=1
\def\buildarxiv{}

\documentclass[10pt,twocolumn,letterpaper]{article}

\usepackage[pagenumbers]{cvpr}

\input{preamble}

\newif\ifmainonly\newif\ifsupponly
\ifdefined\buildmainonly\mainonlytrue\fi
\ifdefined\buildsupponly\supponlytrue\fi
\ifmainonly\usepackage{xr-hyper}\fi
\ifsupponly\usepackage{xr-hyper}\fi

\definecolor{cvprblue}{rgb}{0.21,0.49,0.74}
\usepackage[pagebackref,breaklinks,colorlinks,allcolors=cvprblue]{hyperref}

\def\confName{CVPR}
\def\confYear{2027}

\title{What If This Source Had Never Been Used?\\Condition-Adaptive Diffusion Composition with Exact Source Counterfactuals}

\author{Naoki Murata$^{1}$\quad Yuhta Takida$^{1}$\quad Masaaki Imaizumi$^{2,3}$\\
Toshimitsu Uesaka$^{1}$\quad Chieh-Hsin Lai$^{1}$\quad Yuki Mitsufuji$^{1}$\\[0.6em]
$^{1}$Sony Group Corporation\qquad $^{2}$RIKEN\qquad $^{3}$The University of Tokyo\\[0.4em]
{\small Correspondence to: {\tt naoki.murata@sony.com}}
}

\begin{document}
\ifsupponly\else
\maketitle
\input{sec/0_abstract}
\input{sec/1_intro}
\input{sec/2_related}

\input{sec/3_method}
\input{sec/4_experiments}
\input{sec/5_conclusion}
\clearpage
\input{sec/6_llm_usage}
{
    \small
    \bibliographystyle{ieeenat_fullname}
    \bibliography{main}
}
\fi

\ifmainonly\else
\input{sec/X_suppl}

\ifsupponly
{
    \small
    \bibliographystyle{ieeenat_fullname}
    \bibliography{main}
}
\fi
\fi

\end{document}

%% file: preamble.tex
\usepackage{cuted} %

\usepackage{currfile} %

\usepackage{caption} %

\usepackage{array}

\usepackage{amsthm}
\theoremstyle{definition}
\newtheorem{definition}{Definition}
\theoremstyle{plain}
\newtheorem{proposition}{Proposition}
\newtheorem{corollary}{Corollary}
\AtEndPreamble{
    \crefname{corollary}{Cor.}{Cors.}
    \Crefname{corollary}{Corollary}{Corollaries}
    \crefname{definition}{Def.}{Defs.}
    \Crefname{definition}{Definition}{Definitions}
    \crefname{proposition}{Prop.}{Props.}
    \Crefname{proposition}{Proposition}{Propositions}
}

\usepackage{xspace}

\newcommand{\Gen}{\mathrm{Gen}}          %
\newcommand{\Units}{U}                   %
\newcommand{\cond}[1]{\textup{(C#1)}}    %

\newcommand{\soup}{Relevance Soup\xspace}
\newcommand{\xsoup}{Explain-Away Relevance Soup\xspace}

%% file: sec/0_abstract.tex
\begin{abstract}
How would generated images differ if a source of training data, such as an artist's portfolio, had never been used?
Rebuilding the model without that source answers this exactly but requires expensive retraining for each set of removed sources, while training data attribution methods only approximate the answer.
We instead build diffusion models that answer this question exactly without retraining.
Such a model consists of a shared base model and one component per source, built from its data alone, and removes sources by recombining the remaining components; we call this removal \emph{counterfactually exact} if it yields exactly the generations of the model rebuilt without them.
Equal weights of the components keep exactness but ignore the conditioning input, which harms generation quality, whereas a selector trained on all sources breaks it.
We therefore let each source score from its own component how relevant the input is, and let a rule fixed in advance turn the scores into the weights and the merge of the components.
We show that recombination stays exact whenever the data of each source affect only its own component and the weights and the merge depend only on the conditioning input, the generation state, and the components.
Experiments in text-to-image generation and inpainting show that the models use the sources that each input needs more often than prior compositions that stay exact, and that removing a source changes mainly the generations that relied on it.
Exact counterfactuals thus let us measure how each generation depends on each source, and the composition lets us adjust this dependence.
\end{abstract}

%% file: sec/1_intro.tex
\section{Introduction}
\label{sec:intro}

\begin{figure*}[t]
  \centering
  \includegraphics[width=\linewidth]{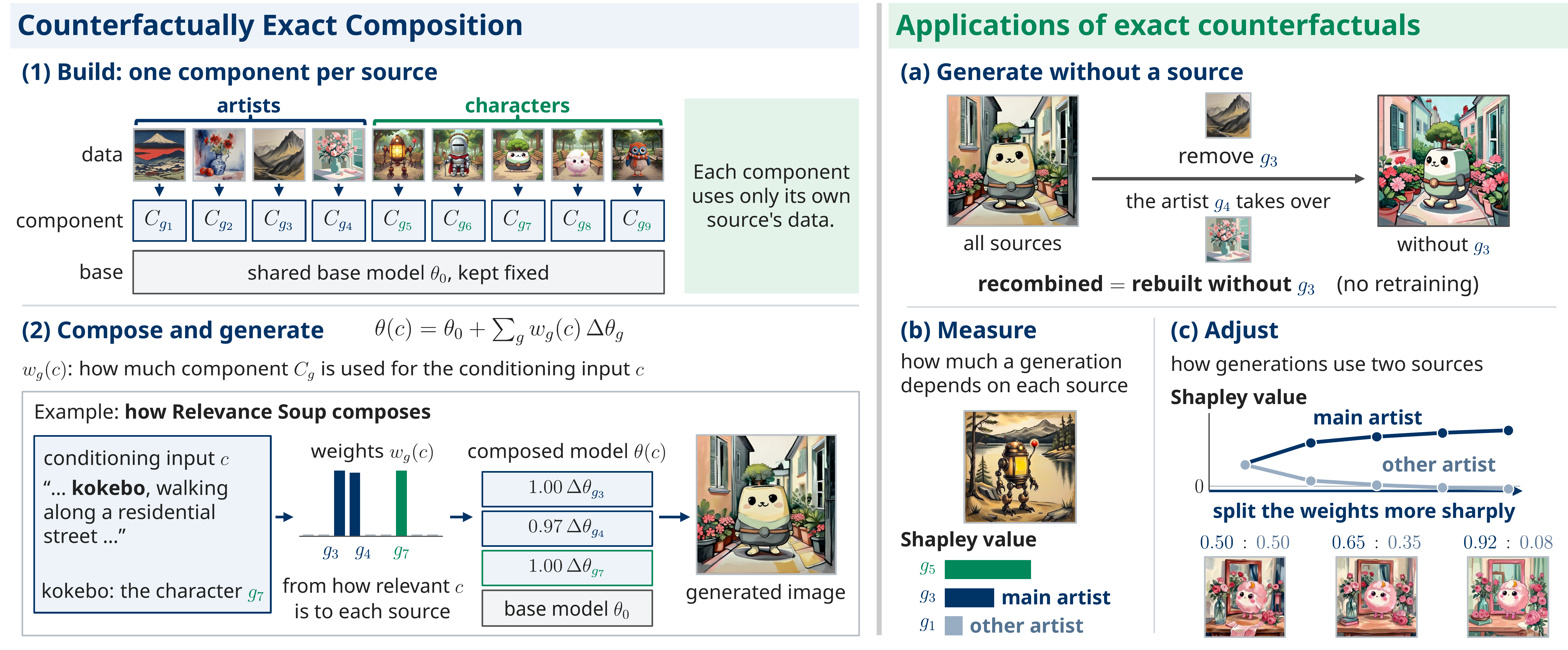}
  \caption{\textbf{Counterfactually exact composition.}
  We build a diffusion model from a fixed base model and one component per source, built from that source's data alone, and compute the weights and the merge of the components from the conditioning input, the generation state, and the available components alone (left; Relevance Soup in the example).
  Recombining the remaining components then gives exactly the model rebuilt without the removed sources, so exact counterfactuals come without retraining: they (a)~show what would have been generated without a source, (b)~measure how much each generation depends on each source, and (c)~let us adjust it through the weights.}
  \label{fig:overview}
\end{figure*}

Conditional diffusion models~\cite{rombach2022ldm} are often adapted to data contributed by many parties, such as a studio's character designs or a manufacturer's product photos~\cite{hu2022lora,ruiz2023dreambooth}.
We refer to the data of each party, the unit in which data are added to or withdrawn from the model, as a \emph{data source}, or simply a \emph{source}.
Once several sources feed a single model, contributors and platforms face counterfactual questions.
An artist may ask how an image would have looked had their portfolio not been used; a rights holder who opts out may ask what the model can still generate without their data; and a platform that compensates contributors needs to know which sources an output depended on~\cite{wang2024economic}.
Answering each of them requires the \emph{counterfactual model}, the model that would have been built without some sources.

Rebuilding the model without the removed sources gives the counterfactual model exactly.
However, each set of sources removed together requires its own rebuild, typically an expensive retraining, and the number of such sets grows rapidly with the number of sources.
Training data attribution methods, based on influence functions~\cite{koh2017influence,park2023trak,zheng2024dtrak,georgiev2023journey,kwon2024datainf,mlodozeniec2025influence}, or unlearning~\cite{wang2024unlearning,serra2026mucs,guda2026} offer a practical and cheaper way to answer these questions.
But they only approximate the counterfactual: their accuracy is not guaranteed for any given output, and most of them return a score rather than the counterfactual model itself.

We instead build models that give the counterfactual model exactly, without retraining.
Such a model consists of a shared pretrained base model and one \emph{component} per source, built independently from that source's data.
Removing sources amounts to recombining the remaining components, and the result is identical to the model rebuilt in the same way without those sources.
Note that, because this exactness holds by construction, our approach applies only to models built from per-source components from the start, not to an existing model trained on all sources together.

Exact counterfactuals let us
(i)~generate, for any set of removed sources, the images that the model would have produced without them (\cref{fig:overview}a), and
(ii)~measure how much each generation depends on each source, for instance which sources are indispensable for it and which can substitute for one another (\cref{fig:overview}b).
Moreover, the way we combine the components also lets us adjust this dependence (\cref{fig:overview}c).

Building such a model poses two challenges.
First, the recombined model, cheap to obtain, must coincide exactly with the rebuilt model, which requires expensive retraining.
Second, this must hold without sacrificing generation quality.
Training each component separately does not by itself address the first challenge.
When combining the components, the model must also decide which components to use and how much, that is, the \emph{weight} of each component, and this decision changes with the \emph{conditioning input}, such as the prompt.
If the weights are produced by a selector trained on all sources~\cite{wu2024mole}, the removed sources still influence the selector, and the recombined model no longer coincides with the rebuilt one.
The simplest way to keep them identical is to give all components equal weights~\cite{biggs2024soup}, but this fails the second challenge.
For example, a prompt that names a character and asks for the style of a particular artist needs the components of that character and of that artist; equal weights dilute them with unrelated ones.

We show that both challenges can be met at once: we define a \emph{composition} as a rule fixed in advance that computes the weights and the merge of the components, design ones that adapt to the conditioning input without sacrificing generation quality (\cref{sec:relevance}), and show that recombination under them is exact (\cref{sec:conditions}).
Instead of a selector trained on all sources, each source scores, from its own component, such as its LoRA and statistics of its data (\eg, embeddings of its captions), how relevant the input is to it.
The components relevant to the input are then combined, and when none is relevant, generation nearly falls back to the base model.
We present two: \emph{Relevance Soup}, which sets the weights from the conditioning input, and \emph{Explain-Away Relevance Soup}, which keeps sources that one part of the input explains from competing for the rest; a merge that drops conflicting updates improves them further.
Recombination is exact because removing a source leaves the other components and the composition unchanged; it suffices that the data of each source affect only its own component, and that the weights and the merge are computed by a function fixed in advance from the conditioning input, the generation state, and the components alone.
These conditions admit many other compositions, suited to each task.

We evaluate these compositions in text-to-image generation, with virtual artists and invented characters as sources, and in inpainting, with door designs and cabinet legs as sources.
In both, they use the needed sources more often than other compositions satisfying the conditions (in text-to-image generation, both sources in up to $64\%$ of images, against at most $26\%$).
Removing a source changes mainly the generations that relied on it, and recombination stays exact, unlike with a trained selector.
Exact counterfactuals measure how each generation depends on the sources, and the composition lets us adjust whether a generation depends mainly on one source or on several.

\noindent In summary, our contributions are:
\begin{enumerate}[leftmargin=*,topsep=2pt,itemsep=2pt]
  \item A framework for exact counterfactuals without retraining: a formal definition of \emph{counterfactual exactness} and sufficient conditions under which recombining the components achieves it.
  \item Compositions within these conditions that adapt the weights to the conditioning input: Relevance Soup and Explain-Away Relevance Soup, which improves on it, both further improved by a sign-aligned merge.
  \item Experiments in text-to-image generation and inpainting showing that these compositions keep generation quality and are exact, unlike common choices outside the conditions, and measure and steer source dependence.
\end{enumerate}

%% file: sec/2_related.tex
\section{Related Work}
\label{sec:related}

\paragraph{Training data attribution.}
Attribution methods for diffusion models approximate how outputs depend on the training data, with influence-function approximations~\cite{zheng2024dtrak,georgiev2023journey,mlodozeniec2025influence} or by unlearning, from the model trained on all data, the generated image~\cite{wang2024unlearning,serra2026mucs} or a group of training data~\cite{guda2026}.
Dai and Gifford~\cite{dai2026unattributable} train an ensemble on coded data splits and drop the members that saw a source, such as one artist's works, which leaves an ensemble smaller than one retrained without it.
Retrieval and style descriptors measure how much an output resembles training images~\cite{wang2023evaluating,somepalli2024csd}, which differs from how it depends on them (\cref{sec:queries}).
Evaluating attribution scores by the linear datamodeling score (LDS)~\cite{ilyas2022datamodels,park2023trak}, valuing data~\cite{ghorbani2019shapley}, and compensating contributors~\cite{wang2024economic} all need models for many subsets of sources, which recombination gives exactly without retraining (\cref{sec:queries}).

\paragraph{Exact removal by design.}
SISA~\cite{bourtoule2021unlearning} trains components on disjoint shards so that a deletion retrains only one shard, and later work combines such components for other purposes (\cref{sec:supp-related}).
In diffusion models, Diffusion Soup~\cite{biggs2024soup} averages models trained on separate shards, which satisfies our conditions with uniform weights, and CDM~\cite{golatkar2023cdm} weights their noise predictions by a classifier over the sources; its class-conditional variant trains this classifier on all sources and thus falls outside our conditions (\cref{sec:within}).
Designing to Forget~\cite{zheng2026designing} and NULLs~\cite{ghosal2026nulls} design models for removal but report results close to, not identical with, those of retraining.
We give conditions under which such designs make removal exact, and compositions within them that adapt to the conditioning input.

\paragraph{Composing diffusion models and adapters.}
Models and adapters are composed by combining their predictions~\cite{liu2022composable,du2023reduce}, by merging fine-tuned parameters~\cite{wortsman2022soups,ilharco2023task}, or by fusing personalized LoRAs, also with weights from the prompt~\cite{tsoumplekas2026prompt} (\cref{sec:supp-related}).
Gates learned over all adapters~\cite{wu2024mole} weight them, and retrieval~\cite{luo2024stylus,zhao2024loraretriever} chooses them, for each input, in Stylus for each part of a prompt.
These methods aim at composition quality rather than counterfactuals, and a selection trained on all sources keeps their influence after a removal (\cref{tab:related} in the supplement).

%% file: sec/3_method.tex
\section{Counterfactually Exact Composition}
\label{sec:method}

This section shows how to build diffusion models whose counterfactuals for removing sources are obtained exactly without retraining.
We define the counterfactual model and exactness (\cref{sec:setup}), derive Relevance Soup and \xsoup toward this goal (\cref{sec:relevance}), and give the conditions under which removing sources from them is exact (\cref{sec:conditions}).

\subsection{Counterfactual models and exactness}
\label{sec:setup}

Each source $g$ holds data $D_g$, such as an artist's portfolio with captions.
For a set $S$ of sources, we write $\mathcal{D}_S=\{(g,D_g)\}_{g\in S}$.
A procedure $\mathcal{A}_H$ that builds a model from data, including any training, gives the model $M_S=\mathcal{A}_H(\mathcal{D}_S;\xi)$; standard training of one model on all data is one such procedure.
Here $H$ collects the settings that are fixed regardless of the set of sources, such as the parameters $\theta_0$ of the pretrained base model, the hyperparameters, reference data prepared independently of all sources, and constants chosen once before any removal.
The randomness $\xi$ of building the model includes the initialization and the data order.
A model $M$ generates an image $\Gen_M(c,z)$ from a conditioning input $c$, such as a prompt or a masked image, and noise $z$, which collects every random number drawn during generation.

For a set $B\subseteq S$ of sources to remove, the \emph{counterfactual model} is the model $M_{S\setminus B}=\mathcal{A}_H(\mathcal{D}_{S\setminus B};\xi)$ that the same procedure builds from the data without $B$: the procedure is run again on $\mathcal{D}_{S\setminus B}$ with the same $\xi$, so whatever depends on the data, such as the contents of the training batches, changes with it.
A \emph{removal procedure} obtains a model without $B$ from the model already built, and the counterfactual model is the target that it should reach.

\begin{definition}[Counterfactual exactness of a removal procedure]
  \label{def:exact}
  For a procedure $\mathcal{A}_H$, a removal procedure $\mathcal{R}_H$ is \emph{counterfactually exact} if, for every set $S$ of sources, all $B\subseteq S$, and all randomness $\xi$, the model $\widetilde M_{S\setminus B}=\mathcal{R}_H(\mathcal{A}_H(\mathcal{D}_S;\xi),B;\mathcal{D}_S,\xi)$ satisfies $\Gen_{\widetilde M_{S\setminus B}}(c,z)=\Gen_{M_{S\setminus B}}(c,z)$ for all conditioning inputs $c$ and all noise $z$, where $M_{S\setminus B}=\mathcal{A}_H(\mathcal{D}_{S\setminus B};\xi)$.
\end{definition}

That is, after $B$ is removed, the model must generate, for every conditioning input and noise, the same image as the model built without $B$ from the start.
Retraining on $\mathcal{D}_{S\setminus B}$ reaches this target by definition, but every set $B$ whose counterfactual we want requires training a new model.
Moreover, exactness by itself does not guarantee generation quality.
Can removal be exact without retraining while the model still generates well?

\subsection{Condition-adaptive composition}
\label{sec:relevance}

We first make removal possible without retraining and then let the weights that combine the sources adapt, so that the model generates well, without breaking exactness.

\paragraph{Per-source LoRAs and recombination.}
A removal needs no retraining if what the remaining sources have learned does not depend on the removed ones.
We therefore keep the base model frozen, so that no source trains it (\cref{fig:overview}), train a LoRA $\Delta\theta_g$ on each $D_g$ alone, and at each step $i$ of generation merge the LoRAs by weights $w^{(i)}_g(c)$,
\begin{equation}
  \theta^{(i)}_S=\theta_0+\sum_{g\in S}w^{(i)}_g(c)\,\Delta\theta_g,
  \label{eq:soup}
\end{equation}
and predict the noise with $\epsilon_{\theta^{(i)}_S}$, where $\epsilon_\theta$ is the guided noise prediction~\cite{ho2022cfg} of the network with parameters $\theta$.
Each step thus needs the network evaluations of the base model, besides computing the weights (\cref{tab:operators}).
Removing a set $B$ then needs no training: \emph{recombination} discards the LoRAs of $B$ and generates again from the same $c$ and $z$ with the remaining ones (\cref{fig:overview}a).
The weights are recomputed for $S\setminus B$, not by zeroing those of $B$, so the remaining weights, and the images, may change.

\paragraph{Weights from per-source relevance.}
To generate well, the weights should adapt to the conditioning input, but adapting them must not break exactness.
Equal weights, $w^{(i)}_g(c)=1/|S|$ for every $c$, as in Diffusion Soup~\cite{biggs2024soup} (\emph{uniform soup}), depend on no $D_g$, so they keep exactness, but ignore the conditioning input.
A selector trained on $\mathcal{D}_S$ adapts the weights, but then the weights of the remaining sources $g\in S\setminus B$ still depend on $\mathcal{D}_B$, so recombination no longer gives the counterfactual model, in which the selector is retrained on $\mathcal{D}_{S\setminus B}$ (\cref{sec:exp-exact}).
We therefore let each source $g$ score, from $D_g$ alone, how relevant the conditioning input $c$ is to it, and let a rule fixed in advance turn the scores of the available sources into weights, that is, into how much each source is used.
Each source thus keeps a \emph{component} $C_g$, its LoRA and statistics of $D_g$, such as embeddings of its captions, from which it scores its relevance.
We call such a rule, which computes the weights and the merge from the input and the components, a \emph{composition}.
Recombination discards the whole component of each removed source; this changes neither the components of the other sources nor the rule, which, as \cref{sec:conditions} shows, suffices for exactness.
We propose two compositions:
\begin{enumerate}[label=\arabic*.,leftmargin=*,topsep=2pt,itemsep=2pt]
\item \textbf{Relevance Soup} computes the weights from the conditioning input alone.
Different parts of a conditioning input are relevant to different sources.
In a prompt, for example, the name of a character and the description of the scene are relevant to separate sources (\cref{fig:overview}), which would compete for a single allocation of weight over the whole prompt.
Relevance Soup therefore splits the conditioning input, by a fixed rule, into units $u\in\Units(c)$, such as the whole prompt and its words.
For each unit, source $g$ computes a relevance $r_g(u)\ge0$, which measures how much better $u$ fits the data of $g$ than reference data.
Relevance Soup normalizes the relevances in each unit together with a constant $w_0>0$,
\begin{equation}
  \pi_{g\mid u}=\frac{r_g(u)}{w_0+\sum_{h\in S}r_h(u)},
  \label{eq:unit-weights}
\end{equation}
and gives each source the largest weight that it obtains in any unit, the same at every step, $w^{(i)}_g(c)=\max_{u\in\Units(c)}\pi_{g\mid u}$.
The base model always enters \cref{eq:soup} with coefficient one, and $w_0$ sets the share of each unit that no LoRA takes.
Hence a unit where every relevance is small compared with $w_0$ nearly falls back to the base model, sources relevant to the same unit share its weight, so that the others take over when one is removed, and sources relevant to different units are used together.
Ideally, only the sources that the input needs receive large weights.
\item \textbf{\xsoup} shows that a composition can be improved without leaving the conditions.
In Relevance Soup, a source that fits a narrow unit, such as the name of a character, also takes a share of the whole prompt; this share never sets its own weight but lowers those of the other sources.
\xsoup leaves such sources and their substitutes (the explained-away set $E$) out of the competition for the whole prompt, so that its weight goes to the sources it still has to explain (\cref{sec:supp-operators}).
\end{enumerate}
\noindent Either composition can also refine the merge, as an option: weighted LoRAs of different sources can push a parameter in opposite directions, and keeping only the terms that agree in sign with their sum in the early steps, where the layout forms (\emph{sign-aligned merge}, after TIES-Merging~\cite{yadav2023ties}), lets them act together; it improves both (\cref{tab:variants}).
Exactness holds for any units, relevance, and weighting rule that read only the components of the available sources (\cref{sec:conditions}), so these can be designed for each task.
In text-to-image generation, the units are the whole prompt, its words, and its pairs of adjacent words, and $r_g(u)$ is a ratio of kernel densities, in the embedding space of a fixed text encoder, of the captions of $g$ and of reference prompts (\cref{sec:supp-relevance}).
In inpainting, the units are the parts of the object, and each source scores the visible region (\cref{sec:exp-inpaint}).

\subsection{Conditions for exact recombination}
\label{sec:conditions}
\label{sec:within}

We now return to exactness and show why recombination under these compositions gives the counterfactual model.
We state the conditions for any model built from per-source components, so that other compositions can be checked against them as well.
Dependence on the data of $B$ could persist in two places: in the components of the remaining sources, if building them used the data of other sources, and in the computation of the weights, if it reads anything built from all sources outside $H$.
\cond{2} rules out the first and \cond{3} the second, while \cond{1} fixes everything that the sources share.

\begin{enumerate}[label=(C\arabic*),leftmargin=*,topsep=2pt,itemsep=2pt]
  \item \textbf{Fixed shared settings.}
  Shared settings are fixed once for all sets of sources: the same $H$ is used for every $S$, including every constant that the composition reads.
  \item \textbf{Source-local construction.}
  Each component is built from its own source alone: $C_g$, such as a LoRA trained on $D_g$, depends only on $D_g$, $H$, and randomness $\xi_g$ derived from $\xi$ and the name of $g$, so it is the same in every model that contains $g$.
  \item \textbf{Generation from the sources in $S$ alone.}
  Generation sees only the components of the sources in $S$: the model generates by a deterministic function $\Phi_H(\{(g,C_g)\}_{g\in S};c,z)$ that is defined for every $S$, including $S=\emptyset$, does not depend on the order of the sources, and reads no other state, such as quantities derived from sources outside $S$.
\end{enumerate}

\begin{proposition}[proof in \cref{sec:supp-proof}]
  \label{prop:exact}
  For a model that satisfies \cond{1}--\cond{3}, recombination is a counterfactually exact removal procedure.
\end{proposition}

Checking \cond{3} directly requires following the whole generation.
For compositions that merge LoRAs by weights, recombination is exact as long as the composition, given only the components of the remaining sources, does the same computation as for any other set of sources.

\begin{corollary}[proof in \cref{sec:supp-proof}]
  \label{cor:adaptive}
  Under \cond{1} and \cond{2}, recombination is counterfactually exact for every composition that, at every step, computes the weights and the merge of the LoRAs from $H$, $c$, $z$, the state of generation, and the components of the sources in $S$ alone, independently of the order of the sources.
\end{corollary}

A natural way to fall outside the conditions is to use the data of other sources when building a component.
For example, if the relevance of a source is learned by a classifier that separates its images from those of the other sources, each component retains the data of the others, violating \cond{2}; restoring exactness would then require retraining the classifiers for every removal, whereas recombination requires no training at all.
Pooled statistics and jointly fitted coefficients also fail (\cref{tab:violations}, bottom).
Comparing the components of the remaining sources during generation, in contrast, is allowed.
\xsoup keeps the two properties that make Relevance Soup exact, a rule fixed in advance and inputs from the sources in $S$ alone: it scores one source on the captions of another to find substitutes, and after a removal it recomputes these scores and the explained-away set among the remaining sources; sign-aligned merge also reads only the weighted LoRAs of $S$.

Exactness thus comes from the structure of \cond{1}--\cond{3}, not from the rule of the composition, so the weights and the merge can be improved without a new proof; all compositions at the top of \cref{tab:violations} satisfy the conditions (\cref{sec:supp-operators}).

\paragraph{Scope of the guarantee.}
Under \cond{1}--\cond{3}, the counterfactual of removing any set $B$ is obtained from the stored components alone, without the data or retraining.
As with retraining, $H$ stays fixed, including the base model, the training settings, and the constants of the composition, such as $w_0$, that we chose once on development data of all sources; choosing them anew from the remaining components by a fixed rule is also exact (\cref{sec:supp-recalibration}).
The conditions are sufficient but not necessary.

\begin{table}[t]
  \centering
  \footnotesize
  \caption{Compositions within the conditions (top) and choices outside them (bottom). $^\dagger$:~tested with our LoRAs (\cref{sec:exp-exact}, \cref{sec:supp-exp-t2i}).}
  \label{tab:violations}
  \setlength{\tabcolsep}{2.5pt}
  \begin{tabular}{@{}ll@{}}
    \toprule
    Composition & Weights computed from \\
    \midrule
    Uniform soup~\cite{biggs2024soup} & nothing (constant) \\
    Stylus-type~\cite{luo2024stylus} & keywords, by a language model \\
    CDM, text-to-image~\cite{golatkar2023cdm} & $\hat x_0$, normalized over the image \\
    Relevance Soup & conditioning input, unit by unit \\
    \xsoup & the above, without explained sources \\
    \quad + sign-aligned merge & the above; merged by sign agreement \\
    \midrule
    Choice outside & Example \\
    \midrule
    One LoRA trained on all data & pooled LoRA (\cref{sec:exp-t2i}) \\
    Selector trained on all data$^\dagger$ & class-cond.\ CDM~\cite{golatkar2023cdm}, gates~\cite{wu2024mole} \\
    Statistics pooled over sources$^\dagger$ & prior of all experts~\cite{li2022btm} \\
    Coefficients fitted jointly$^\dagger$ & ZipLoRA~\cite{shah2024ziplora} \\
    \bottomrule
  \end{tabular}
\end{table}

%% file: sec/4_experiments.tex
\section{Experiments}
\label{sec:experiments}

We ask three questions.
First, can compositions within our conditions generate well, using the sources a conditioning input needs and falling back to the base model otherwise, and can they be improved further without leaving the conditions (\cref{sec:exp-t2i,sec:exp-inpaint})?
Second, is recombination exact in practice, and do the conditions matter, or would separately built components suffice (\cref{sec:exp-exact})?
Third, what do the exact counterfactuals reveal about how each generation depends on the sources, and can the composition steer this dependence (\cref{sec:exp-attribution})?

\subsection{Text-to-image generation}
\label{sec:exp-t2i}

\begin{table}[t]
  \centering
  \small
  \caption{Text-to-image results. Left: prompts that name a character in a covered subject ($720$ generations), where a composition should use the sources; joint success needs both identity and a covering artist's style, a similarity criterion that also accepts some base images, and CLIP replaces the name by ``a character''. Right: prompts that no source fits, where it should stay close to the base model; Leak.\ is the largest CSD share of one artist on uncovered subjects, and FID uses $5{,}000$ COCO captions (\cref{sec:supp-exp-t2i}). The base model and the pooled LoRA are references; bold: best composition under the conditions (left). Indented rows add to the row above (\soup + explain-away $=$ \xsoup).}
  \label{tab:t2i}
  \setlength{\tabcolsep}{1.5pt}
  \begin{tabular}{@{}lcccccc@{}}
    \toprule
    & \multicolumn{4}{c}{Named character} & \multicolumn{2}{c}{No source fits} \\
    \cmidrule(lr){2-5}\cmidrule(l){6-7}
    Method & Joint$\uparrow$ & ID$\uparrow$ & Style$\uparrow$ & CLIP & Leak.$\downarrow$ & FID$\downarrow$ \\
    \midrule
    SDXL base~\cite{podell2024sdxl} & 0.00 & 0.00 & 0.29 & 29.9 & 0.25 & 24.7 \\
    \midrule
    Uniform soup~\cite{biggs2024soup} & 0.00 & 0.00 & 0.34 & 30.3 & 0.22 & 31.5 \\
    CDM, text-to-image~\cite{golatkar2023cdm} & 0.00 & 0.00 & \textbf{0.92} & 26.8 & 0.43 & 37.4 \\
    Stylus-type~\cite{luo2024stylus} & 0.26 & 0.48 & 0.65 & 25.5 & 0.84 & 27.9 \\
    \multicolumn{7}{@{}l}{\emph{Ours}} \\
    \soup & 0.50 & 0.68 & 0.78 & 24.4 & 0.26 & 25.5 \\
    \enspace + explain-away & 0.57 & 0.64 & 0.88 & 24.5 & 0.26 & 25.5 \\
    \quad\enspace + sign-aligned merge & \textbf{0.64} & \textbf{0.76} & 0.85 & 23.9 & 0.26 & 25.6 \\
    \midrule
    Pooled (reference) & 0.21 & 0.56 & 0.51 & 25.2 & 0.50 & 34.7 \\
    \bottomrule
  \end{tabular}
\end{table}

\begin{figure*}[t]
  \centering
  \includegraphics[width=\linewidth]{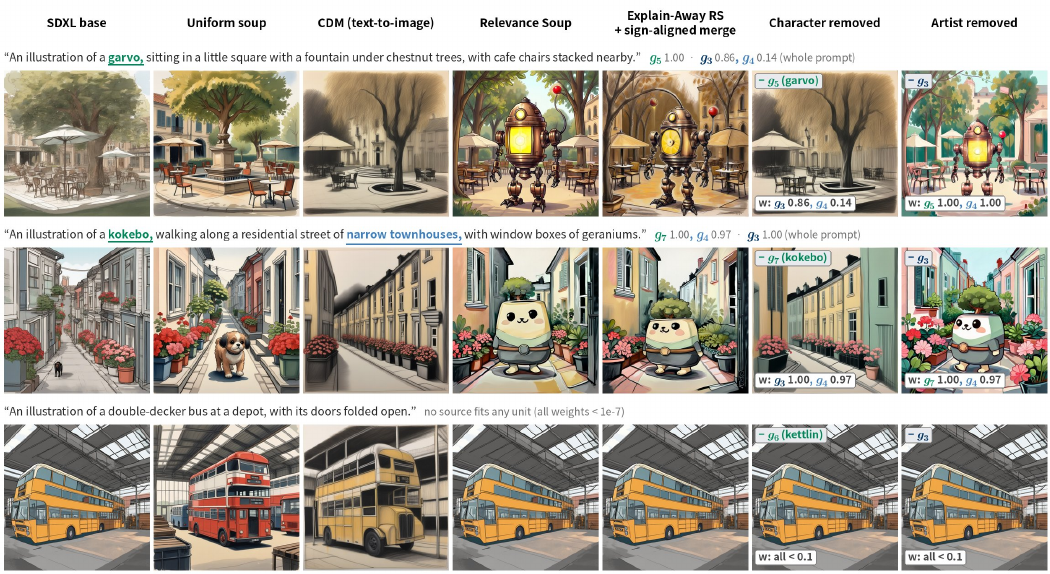}
  \caption{\textbf{Text-to-image generations and removals.}
  Top and middle: a named character in a subject covered by two artists; bottom: a subject that no source covers.
  Uniform soup draws neither the character nor an artist's style, and CDM draws only an artist's style, even where no source fits.
  Prompts show the weights of \xsoup (Explain-Away RS) with sign-aligned merge, from which the removals are made, and boxes the weights after each removal.
  In the top row, Relevance Soup gives the covering artist $g_3$ a weight of only $0.17$ and misses its style; Explain-Away RS raises the weight to $0.86$.
  Removing the character removes it, and removing a covering artist leaves the character in the style of the other ($g_4$ without $g_3$).
  For the uncovered subject, our compositions give the image of the base model, which removals leave unchanged.}
  \label{fig:t2i-qualitative}
\end{figure*}

\paragraph{Setup.}
We build nine sources on SDXL~\cite{podell2024sdxl}, each a LoRA trained on its own synthetic images: four virtual artists $g_1$--$g_4$ with distinct techniques and palettes, each covering three of six subjects (such as mountains and still lifes) so that each covered subject has two artists, and five invented characters $g_5$--$g_9$.
Synthetic sources tell us which sources each prompt needs and use no artist's work without consent.
The task is built so that a prompt needs two sources at once, as when a request asks for one contributor's character in the domain of another.
The artists are never named and must be inferred from new scene descriptions, and the ranking holds on paraphrased prompts (\cref{sec:supp-exp-t2i}).

The main measure is joint success: the image shows the named character, as decided by a Qwen3-VL judge~\cite{bai2025qwen3vl}, in the style of an artist that covers the subject, as measured with CSD~\cite{somepalli2024csd}.
The judge is checked on control images and against a judge of another family, and the judges of both tasks agree with one author's blind judgments on an estimated $0.85$ and $0.84$ of outputs (\cref{tab:human}).
We compare ours with three prior compositions under the conditions and a LoRA trained on all data (pooled).

\paragraph{Our compositions use the sources that a prompt needs.}
On prompts that name a character in a covered subject, ours draw the named character in the style of a covering artist more often than the other compositions under the conditions: joint success is $0.50$ for \soup and $0.64$ with \xsoup and sign-aligned merge (\cref{tab:t2i}, \cref{fig:t2i-qualitative}).
Because our weights are normalized within each part of the prompt, the character and the covering artists receive large weights from different units.
\soup thus shows both more often than the pooled LoRA, which learns all sources in one adapter.
Uniform soup scales every LoRA down too far to draw either, and CDM puts its weight on one artist, drawing only its style.
The Stylus-type selection also assigns sources to parts of the prompt but succeeds less often than ours, so the gain comes from how each source scores the parts from its own component, not from splitting the prompt alone.

\paragraph{Prompts that no source fits stay close to the base model.}
On prompts that name no character in a subject that no artist covers, and on COCO~\cite{lin2014coco} captions, our compositions nearly fall back to the base model: their artist style and FID~\cite{heusel2017fid} stay close to those of the base model, whereas the others impose artist style or raise the FID.

\paragraph{The composition can be improved within the conditions.}
\xsoup raises the style rate and joint success, because the name of the character no longer takes a share of the whole prompt from the covering artists; identity falls slightly.
Adding sign-aligned merge recovers identity and raises joint success by $0.135$ over \soup (95\% interval $[0.08,0.19]$). It needs the same network evaluations and $1.17$ times the time, and lowers CLIPScore by $0.5$. Every method that draws the character has a lower CLIPScore than the base model, partly as it draws the scene less often; on COCO captions, \soup changes it by $-0.13$ (\cref{sec:supp-exp-t2i}).
With sources trained from two other seeds and with another judge, both still exceed \soup (\cref{sec:supp-exp-t2i}), and with ten more characters ($19$ sources), \soup keeps its joint success ($0.49$) and exact removal (\cref{tab:scaling}).

\subsection{Inpainting}
\label{sec:exp-inpaint}

To test the first question with a condition given as an image rather than text, we turn to inpainting, where the visible region calls for different sources for different parts.

\paragraph{Setup.}
Three sources specify parts of a cabinet on the SDXL inpainting U-Net~\cite{diffusers2023sdxlinpaint}: two door designs, a diamond medallion under an arched frame ($g_1$) and a sunburst under a stepped frame ($g_2$), and brass legs with corner caps ($g_3$); each LoRA is trained to follow its data only within its part (\cref{sec:supp-relevance}).
The doors and legs are masked; only the top band is visible, whose frame and corner caps cue the sources.
Test images need a door design and the legs, one of them, or neither, and a fill should use exactly the cued sources, or the base model when none is cued.
Each source scores the visible region with a logistic classifier on CLIP features~\cite{radford2021clip} fitted to separate its own images from fixed generic cabinets, and we use \soup with the parts as units, keeping in each part only the most relevant of the base model and its sources.
We compare it with the base model, uniform soup, and an oracle that uses the needed sources, on $396$ test images, $300$ of which need two sources (\cref{tab:inpaint}).
A Qwen3-VL judge and a second rater check whether a fill shows the required door and legs (\cref{sec:supp-exp-inpaint}).

\begin{figure}[t]
  \centering
  \includegraphics[width=\linewidth]{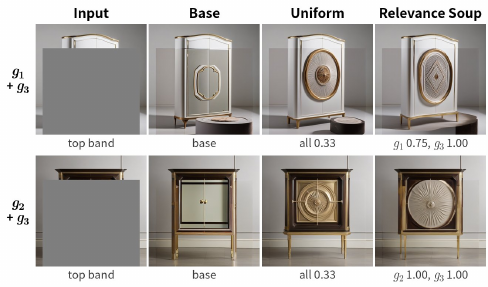}
  \caption{\textbf{Inpainting from the top band.} The input is only the top band, whose frame cues a door design ($g_1$ or $g_2$) and whose corner caps cue the brass legs ($g_3$); labels give the weights. The base model ignores the cues and uniform soup imposes every source; Relevance Soup uses the cued door and the legs.}
  \label{fig:inpaint-main}
\end{figure}

\begin{table}[t]
  \centering
  \small
  \caption{Inpainting success on $396$ test images with two seeds, averaged over the six kinds (All), the kinds that need two sources (Two), and those that need at most one (One/none); a composition should be high on Two and not below the base model on One/none. Second rater: another vision-language model shown no method names. The base model and the oracle, which uses the needed sources given by hand, are references; bold: best composition under the conditions.}
  \label{tab:inpaint}
  \setlength{\tabcolsep}{3.2pt}
  \begin{tabular}{@{}lccccc@{}}
    \toprule
    & \multicolumn{3}{c}{Qwen3-VL judge} & \multicolumn{2}{c}{Second rater} \\
    \cmidrule(lr){2-4}\cmidrule(l){5-6}
    Method & All & Two & One/none & All & Two \\
    \midrule
    Inpainting base & 0.30 & 0.05 & 0.42 & 0.14 & 0.01 \\
    \midrule
    Uniform soup~\cite{biggs2024soup} & 0.26 & 0.36 & 0.21 & 0.20 & 0.25 \\
    \soup & \textbf{0.45} & \textbf{0.47} & \textbf{0.45} & \textbf{0.34} & \textbf{0.39} \\
    \midrule
    Needed sources (oracle) & 0.83 & 0.79 & 0.84 & 0.62 & 0.63 \\
    \bottomrule
  \end{tabular}
\end{table}

\paragraph{Relevance Soup combines the sources that an image needs.}
On the images that need two sources, Relevance Soup succeeds far more often than the base model (\cref{tab:inpaint}, \cref{fig:inpaint-main}).
Uniform soup also uses the sources together but imposes all of them on images that need fewer, so it falls below Relevance Soup on both groups with both raters (One/none, second rater: $0.18$ vs.\ $0.32$).
On One/none, Relevance Soup is only on par with the base model, as it also brings a source into images that need none (\cref{sec:supp-exp-inpaint}).

\subsection{Exactness}
\label{sec:exp-exact}

\paragraph{Recombination is exact in practice.}
\Cref{prop:exact} holds in exact arithmetic; in floating point, it also needs the same computations in the same order (\cref{sec:supp-proof}).
In every case we tested, in both tasks, including $72$ of $72$ removals for each of seven text-to-image compositions, recombination was bit-identical to a model built in a separate process from the components of the remaining sources (\cref{sec:supp-exp}).

\paragraph{Separate components alone do not make recombination exact.}
We keep the same separately trained LoRAs but compute the weights with a choice outside the conditions (\cref{tab:violations}, bottom).
With a selector trained jointly on all sources, as in class-conditional CDM, recombination keeps the selector, whereas the counterfactual model retrains it without the removed source.
Its weights then differ from those of the counterfactual model in every generation, even when the removed source is unrelated to the prompt, where the image mismatch exceeds the change that the removal itself causes in our compositions; the other choices that we tested also break exactness (\cref{sec:supp-exp-t2i}).

\subsection{Measuring and steering source dependence}
\label{sec:exp-attribution}

Because recombination gives every subset model exactly, how a generation depends on its sources can be measured directly (\cref{fig:overview}b), from the regenerated images and from the denoising loss of a fixed image when sources are removed (\cref{sec:queries}).
This differs from the composition weights, which measure how much a source is used: after a removal, other sources can take over.

\paragraph{Removing a source changes mainly the generations that relied on it.}
Removing the named character removes it: with each of our compositions, its identity rate falls to $0.00$.
When a covering artist is removed, the other artist that covers the subject takes over in most generations, which its weight before the removal does not show (\cref{fig:t2i-qualitative}, \cref{sec:supp-exp-t2i}).
Removing a source unrelated to the prompt changes the images of our compositions by a DINOv2~\cite{oquab2024dinov2} distance of at most $0.025$, against $0.07$--$0.15$ for other compositions and $0.17$ between noise seeds (\cref{sec:supp-exp-t2i}).

\paragraph{Single removals miss redundant sources.}
We add to the nine text-to-image sources two near-duplicates (twins), $g_{10}$ of $g_5$ and $g_{11}$ of $g_7$.
Removing either twin leaves the character in the regenerated images of \soup, and only removing both removes it; removing one twin raises the loss of the generated image by less than half as much as removing both (\cref{sec:queries}).
A source can thus be needed even when removing it alone changes little, which single-removal attribution misses~\cite{ghorbani2019shapley}.

\begin{figure}[t]
  \centering
  \includegraphics[width=\linewidth]{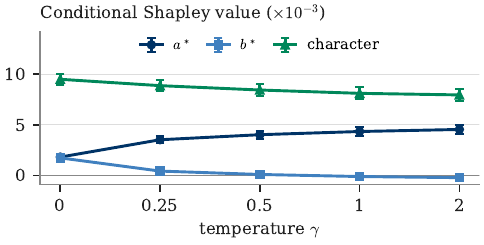}
  \caption{\textbf{Steering source dependence by the composition rule.} Conditional Shapley values ($\times10^{-3}$) of the covering artist with the larger weight under \soup ($a^*$), the other ($b^*$), and the character against the temperature $\gamma$ ($0$: equal split; $1$: \soup; ticks evenly spaced); means and $95\%$ intervals over $238$ generations (joint success and identity: \cref{fig:steering}).}
  \label{fig:steering-main}
\end{figure}

\paragraph{The composition rule can steer the dependence.}
With the components fixed, we split the weight that \soup gives the two artists covering the subject of a prompt in proportion to their scores raised to a temperature $\gamma$, keeping its total ($\gamma=0$ splits it equally, $\gamma=1$ is \soup).
For each $\gamma$, we compute conditional Shapley values~\cite{shapley1953value,ghorbani2019shapley} of the two artists and the character from exact subset losses (\cref{sec:supp-steering}).
At $\gamma=0$, the two artists receive the same credit on average; as $\gamma$ grows, the credit moves on average monotonically to the artist with the larger weight, with little change beyond $\gamma=2$ (\cref{fig:steering-main}, \cref{fig:steering}).
Spreading the dependence has a cost: at $\gamma=0$, the style of the covering artists weakens and joint success falls, while identity rises.
The value of the character also falls as $\gamma$ grows, although its weight stays near one, a shift that the weights alone do not show.
The rule thus lets a platform choose whether a generation concentrates its dependence on one source or spreads it over several, and verify it exactly.

%% file: sec/5_conclusion.tex
\section{Conclusion}
\label{sec:conclusion}

Exact source counterfactuals need not come from retraining.
When the data of each source affect only its own component and a fixed rule computes the weights and the merge from the conditioning input, the generation state, and the available components alone, recombination is exact.
Within these conditions, the weights can adapt to the conditioning input and the merge can be refined, so one model generates well, shows how each generation depends on each source without retraining, and lets us adjust this dependence.

\paragraph{Limitations and outlook.}
The guarantee is relative to our construction and does not cover the data used to pretrain the base model or a model trained jointly on all sources.
Our experiments use synthetic sources in two tasks, text-to-image generation and inpainting; other settings may admit better compositions, which remain exact as long as they satisfy the conditions.
The natural next step is the portfolios of consenting artists, where exact counterfactuals can answer the questions of contributors and platforms.

%% file: sec/6_llm_usage.tex
\section*{LLM Usage Statement}

\noindent\textit{Research.}
As described in \cref{sec:supp-exp}, generative models serve as tools in our pipeline: SDXL generated the images of the artists and the cabinets and the reference images of the characters, from which Qwen-Image-Edit generated the character images; Qwen3-VL-8B, Phi-4, Gemma~3, and Claude Sonnet filtered data or judged generations; Qwen3-VL-8B also served as the selector of the Stylus-type baseline and detected the part regions for inpainting; an LLM-based agent drafted part of the evaluation phrases, and an AI agent visually checked the inpainting test images.

\noindent\textit{Code.}
Most of the experiment code, job scripts, analysis, and figure scripts were written with AI coding assistants (Claude Code with Claude models, and OpenAI Codex) under the authors' direction; the authors reviewed the code and the reported results.

\noindent\textit{Writing.}
The authors determined the content, claims, and structure of the paper.
Following their instructions and edits, Claude was used to draft and revise the English wording and for editing and grammar; the authors reviewed and revised every passage.

%% file: sec/X_suppl.tex
\clearpage
\newcommand{\ifarxivelse}[2]{\ifdefined\buildarxiv#1\else#2\fi}
\ifdefined\buildarxiv\else\setcounter{page}{1}\fi
\ifsupponly\else\ifdefined\buildarxiv\else\renewcommand{\thepage}{\texorpdfstring{\arabic{page}}{S\arabic{page}}}\fi\fi
\maketitlesupplementary
\appendix
\crefname{appendix}{Sec.}{Secs.}\Crefname{appendix}{Section}{Sections}
\setcounter{figure}{0}\renewcommand{\thefigure}{S\arabic{figure}}
\setcounter{table}{0}\renewcommand{\thetable}{S\arabic{table}}
\setcounter{equation}{0}\renewcommand{\theequation}{S\arabic{equation}}

\noindent\textbf{Contents.} \Cref{sec:supp-related}: related work; \cref{sec:supp-proof}: proofs and the scope of the guarantee; \cref{sec:supp-relevance}: relevance scores; \cref{sec:supp-operators}: the compositions; \cref{sec:queries}: counterfactual queries; \cref{sec:supp-steering}: steering source dependence; \cref{sec:supp-exp}: experimental details for text-to-image (\cref{sec:supp-exp-t2i}) and inpainting (\cref{sec:supp-exp-inpaint}); \cref{sec:supp-recalibration}: removal with recalibration; \cref{sec:supp-deletion}: recombination versus component deletion; \cref{sec:supp-baselines}: further baselines and ablations.

\section{Extended Related Work}
\label{sec:supp-related}

This section covers work that \cref{sec:related} mentions only briefly.

\paragraph{Training data attribution and data valuation.}
Influence functions~\cite{koh2017influence} estimate the effect of removing a training example by linearizing training around the final model, and DataInf~\cite{kwon2024datainf} makes this efficient for LoRA-tuned models.
Wang \etal~\cite{wang2024unlearning} unlearn a synthesized image and attribute it to the training images that the model forgets along with it.
DAS~\cite{lin2025das} scores a training image by the change in the predicted distribution when the image is removed, computed by linearizing the noise predictor, and Wang \etal~\cite{wang2025fastgda} distill an improved variant of the attribution of~\cite{wang2024unlearning} into a feature embedding in which influential training images are retrieved in seconds; both estimate how the model trained on all data depends on individual images, whereas recombination answers removal queries exactly at the level of sources.
Wang \etal~\cite{wang2023evaluating} benchmark attribution with images from models customized toward exemplar images and tune features to retrieve those exemplars, Somepalli \etal~\cite{somepalli2023forgery} retrieve training images that generations replicate, and CSD~\cite{somepalli2024csd} compares styles; all of them score how closely an output resembles training images, which, as \cref{sec:related} notes, need not reflect how it depends on them.
Attribution also underlies proposals to value and compensate data contributors~\cite{ghorbani2019shapley,wang2024economic}.
Lin \etal~\cite{lin2025crediting} credit data contributors for global properties of a diffusion model, such as image quality or demographic diversity, with Shapley values, and approximate each model retrained on a subset of contributors by fine-tuning on that subset a pruned model first fine-tuned on all data; for a model built under our conditions with contributors as sources, recombination gives its subset models exactly.
Morreale \etal~\cite{morreale2025attribution} motivate attribution by compensating artists but set aside attribution to training data as intractable, attributing a generation to its inference-time references instead; our models make it exact at the level of data sources.

\paragraph{Exact removal from separately trained components.}
Diffusion Soup~\cite{biggs2024soup} unlearns a shard by averaging the parameters of the models trained on the other shards; with uniform weights and an initialization that excludes the shard data, it satisfies our conditions.
CDM~\cite{golatkar2023cdm} sums the noise predictions of models trained on separate sources, weighted by a classifier over the sources.
Its class-conditional variant trains this classifier on all sources and reads it at generation, which falls outside \cond{3} (\cref{tab:violations}), whereas its text-to-image variant compares the CLIP embedding of the predicted clean image with embeddings of each source's data and satisfies them as long as the clean image is not predicted with the removed sources.
SAFE~\cite{dukler2023safe} lets the component of each shard also use data from a few connected shards, so that a deletion retrains only the components that used the deleted data.
S3T~\cite{chowdhury2025s3t} trains the LoRA layers of a model one after another, the $i$-th on the first $i$ slices of a shard, and, on a deletion, deactivates the layer that first used the deleted slice and all later ones; like dropping ensemble members~\cite{dai2026unattributable}, this leaves a model that does not depend on the deleted data but is trained on less than the remaining data.
\`A-la-carte prompt tuning~\cite{bowman2023alacarte} trains prompts on separate data subsets and combines those that a user may access.
CP-$\Delta$~\cite{vyas2023provable} combines models trained on separate shards to bound the risk of reproducing protected content.
Language models follow the same pattern with nonparametric datastores~\cite{min2024silo}, independently trained domain experts~\cite{li2022btm}, and experts that data owners train on a shared public model~\cite{shi2025flexolmo}.
In FlexOLMo, the router embedding of each expert is trained on its owner's data.
In our notation, $H$ comprises the public model, its expert and its fixed router embedding, the encoders, and the routing rule; $C_g$ comprises the expert of owner $g$, its router embedding, and its bias; and $\Phi_H$ recomputes the routing at every layer and token from the remaining experts.
Interactions between experts during inference are permitted, since they are recomputed from the remaining set.
The reported main model, however, additionally tunes the router after merging on public proxy data selected for each owner.
The tuned router then depends on the owners' data, so our conditions no longer cover the removal of a single expert, although we have not measured whether leaving out an expert then differs from the model built without it.
SIFT-Masks~\cite{kuo2025siftmasks} fine-tunes a language model separately on each task with updates constrained to a random sign vector drawn before training, sums the task vectors, and serves each task through a mask computed from its own vector; with the pretrained model and the sign vector in $H$, the published design satisfies our conditions.

\paragraph{Unlearning in diffusion models.}
Data unlearning fine-tunes a trained diffusion model to remove specific training images or classes~\cite{fan2024salun,alberti2025siss,wu2025erasediff}; methods that target the model retrained without them~\cite{fan2024salun,alberti2025siss} approximate it, whereas recombination produces that model exactly.
Concept erasure fine-tunes a diffusion model to suppress what a prompt describes~\cite{gandikota2023erasing,kumari2023ablating}.
Unlike data unlearning, it does not target the model trained without any particular data and therefore addresses a different question from ours.

\paragraph{Composition of diffusion models and adapters.}
SuperDiff~\cite{skreta2025superdiff} mixes or intersects pretrained models with density estimates computed during sampling, and TIES-Merging~\cite{yadav2023ties} resolves interference when merging fine-tuned parameters.
For personalization by fine-tuning~\cite{ruiz2023dreambooth}, often with LoRAs~\cite{hu2022lora}, multi-concept methods fuse LoRAs by optimization~\cite{gu2023mixofshow,shah2024ziplora}, train them to be summed with little interference~\cite{po2024orthogonal,lee2024dco}, switch or combine them across denoising steps~\cite{zhong2024multilora}, choose between them per attention layer~\cite{ouyang2025klora}, or assign them to image regions or tokens~\cite{yang2025loracomposer,tara2026}.
Like \soup, prompt-aware weighting~\cite{tsoumplekas2026prompt} weights each LoRA by the prompt, through the text-encoder similarity between the prompt and its trigger words or the change of the prompt embedding when they are removed, normalized over the LoRAs in use; since each score reads only the prompt and that LoRA's trigger words, this weighting satisfies our conditions when the LoRAs in use are the available sources and, for its switching variant, their order is fixed by rule.
It assumes that the LoRAs a prompt needs are given and scores the whole prompt, whereas \soup selects them by scoring units of the input and falls back to the base model when none fits.
ZipLoRA and K-LoRA target the combination of a subject and a style, as in our compositions of a character and an artist, but leave the choice of the two adapters to the user.
These methods aim to compose concepts well rather than to answer counterfactual questions; adapted to meet our conditions, they could combine their generation quality with exact counterfactuals.
Our conditions apply alike to these designs, to parameter soups, and to compositions of noise predictions, and they also constrain reference statistics, calibration, and random numbers.

\paragraph{Automatic selection of sources.}
Stylus~\cite{luo2024stylus} retrieves adapters for a prompt from a large collection using a description of each adapter, after which a language model reads the prompt together with the candidates and assigns adapters to the keywords of the prompt.
LoraRetriever~\cite{zhao2024loraretriever} represents each LoRA by the embeddings of a few of its training examples and retrieves the closest LoRAs for each input, using a retriever tuned on a subset of the tasks.
Both methods thus select adapters per input, as do mixture-of-LoRA methods, which learn gates over the adapters~\cite{wu2024mole}; when such gates are trained with every adapter in place, leaving out an adapter need not yield the pipeline built without it, and reading such a shared gate falls outside \cond{3}.
A choice that reads the remaining candidates jointly is not a selector trained on all sources and can satisfy our conditions.
For Stylus, this requires a fixed base model, vision-language model, embedding model, and language model, a description of each adapter derived from that adapter alone, retrieval and assignment re-run on the remaining candidates, and an order and random numbers fixed by rule.
Our adaptation meets these requirements (\cref{sec:supp-operators}); because the reported pipeline draws random masks and samples its language model at a nonzero temperature without specifying how these random numbers are drawn, and we have not inspected its implementation, whether the original satisfies them is not established, and since our conditions are sufficient, falling outside them would not by itself show that recombination is inexact.
A retriever tuned on the data of some sources, in contrast, is shared and changes the embeddings of all remaining LoRAs when it is retuned without a removed one, so removing such a source falls outside \cond{3}.
Our compositions instead rely on relevance scores that each source computes independently.
In language models, BTM~\cite{li2022btm} weights its experts by a domain posterior computed from each expert's likelihood of the context, which is a source-local score.
Its prior, however, is an average of the posteriors of all experts on development data that is cached and fixed at test time, so leaving out an expert falls outside \cond{3}; experts added in later rounds are, moreover, initialized from parameter averages of existing ones, outside \cond{2}.
Two routers for language models read each adapter only through quantities built from it: Arrow~\cite{ostapenko2024arrow} takes the top right singular vector of each LoRA's update at each layer as that LoRA's prototype without using data, and PHATGOOSE~\cite{muqeeth2024phatgoose} trains a gate vector for each expert on that expert's dataset after the expert is trained, with all other parameters fixed.
Both route each token at each layer by a top-$k$ softmax over the experts present, so they satisfy our conditions for experts trained on one source each from a fixed base, whereas the library that Arrow's paper builds by clustering tasks on the parameters of all their adapters does not, since each of its adapters is trained on the joined data of a cluster of tasks and a removal can change the clusters.
\Cref{tab:related} summarizes the designs above.

\begin{table*}[t]
  \centering
  \small
  \caption{How composition methods select their sources. ``Local'' means that the selection reads each source only through quantities built from that source alone. ``No'' means that the design falls outside our sufficient conditions, not that it has no exact removal procedure; ``not established'' means that the reported pipeline does not specify enough to decide whether it satisfies them. Entries analyze the published designs, not their implementations.}
  \label{tab:related}
  \setlength{\tabcolsep}{2pt}
  \begin{tabular}{@{}lll@{}}
    \toprule
    Method & Selection adapts to & Satisfies our conditions \\
    \midrule
    Diffusion Soup~\cite{biggs2024soup} & nothing (uniform) & yes \\
    CDM, class-cond.~\cite{golatkar2023cdm} & noisy image (learned, joint) & no, \cond{3} \\
    CDM, text-to-image~\cite{golatkar2023cdm} & predicted image (local) & yes, if $\hat x_0$ excludes $B$ \\
    MoLE~\cite{wu2024mole} & input (learned, joint) & no, \cond{3} \\
    Stylus~\cite{luo2024stylus} & prompt keywords (LLM) & not established; yes as adapted \\
    LoraRetriever~\cite{zhao2024loraretriever} & input (tuned retriever) & no if a removed source tuned the retriever, \cond{3} \\
    BTM~\cite{li2022btm} & context (local, pooled prior) & no, \cond{3} \\
    FlexOLMo~\cite{shi2025flexolmo} & token (local) & yes, without router tuning \\
    Ours & parts of the condition (local) & yes (\cref{cor:adaptive}, \cref{sec:supp-operators}) \\
    \bottomrule
  \end{tabular}
\end{table*}

\section{Proofs and Scope of the Guarantee}
\label{sec:supp-proof}

\begin{proof}[Proof of \cref{prop:exact}]
  Fix a set $S$ of sources, $B\subseteq S$, the randomness $\xi$, a condition $c$, and noise $z$, and write $C_{g\mid S}$ for the component of $g$ in the model built on $S$.
  By \cond{1}, the constructions on $S$ and on $S\setminus B$ use the same $H$.
  For every $g\in S\setminus B$, the data $D_g$ and the randomness $\xi_g$ do not depend on the set being built, because $\xi_g$ is determined by $\xi$ and the identity of $g$; by \cond{2}, $C_{g\mid S\setminus B}=C_{g\mid S}$. In particular, the training batches of $g$ are drawn from $D_g$ with $\xi_g$ and do not change when other sources are removed.
  Recombination and the counterfactual model therefore give $\Phi_H$ the same set $\{(g,C_g)\}_{g\in S\setminus B}$, together with the same $c$ and $z$, and by \cond{3} $\Phi_H$ is a deterministic function of these arguments alone.
  Hence $\Gen_{\widetilde M_{S\setminus B}}(c,z)=\Gen_{M_{S\setminus B}}(c,z)$.
\end{proof}

\begin{proof}[Proof of \cref{cor:adaptive}]
  Write the weights and the merge of the composition at step $i$ as
  \begin{equation}
    \begin{split}
    w^{(i)}_g&=f\bigl(g,c,z,i,x_{<i},\{(h,C_h)\}_{h\in S}\bigr),\qquad g\in S,\\
    \theta^{(i)}_S&=\theta_0+m_i\bigl(\{(w^{(i)}_g,\Delta\theta_g)\}_{g\in S}\bigr),
    \end{split}
    \label{eq:adaptive}
  \end{equation}
  where $x_{<i}$ is the state of generation before step $i$, and $f$ and $m_i$ are fixed by $H$, defined for every $S$, and invariant to the order of the sources; in \cref{eq:soup}, $m_i$ is the weighted sum $\sum_{g\in S}w^{(i)}_g\Delta\theta_g$.
  At every step, the state, the weights, and the merged parameters are computed only from $H$, $c$, $z$, and $\{(g,C_g)\}_{g\in S}$, and after a removal they are computed anew for the remaining sources; hence \cond{3} holds and \cref{prop:exact} applies.
\end{proof}

\paragraph{What the guarantee compares.}
The proposition equates the generated images for every $(c,z)$, with the randomness of the construction and of sampling matched between the two sides.
This is stronger than equality in distribution, which would only require the two models to generate the same distribution of images, and it does not state that the parameters or internal states of the two models agree beyond what the composition reads.
In floating point, the equality holds bit for bit when both sides run the same computations in the same order, with the same implementation and deterministic kernels; our checks in \cref{sec:supp-exp} confirm it for the cases tested, not for all inputs.
With $H$ fixed, these checks compare recombination with a separate process that loads only the stored components of the remaining sources, so they test the composition, the stage at which the choices outside the conditions in \cref{sec:supp-exp-t2i,sec:supp-deletion} differ from their counterfactual models; that the components themselves are source-local (\cond{2}) follows from how each is built, from $D_g$, $H$, and $\xi_g$ alone.
Matching the randomness is what lets the two sides be compared image by image: a model rebuilt with other random numbers, such as sources trained from another seed, generates different images even without a removal, so a comparison without matched randomness could not separate the effect of the removal from that of the randomness.

\paragraph{Random numbers in the composition.}
Because $z$ collects every random number drawn during sampling, \cond{3} already fixes how the composition uses randomness: recombination must draw its random numbers for the remaining set as the counterfactual model does.
Two implementations satisfy this.
Rerunning the composition on the remaining set from the start of the same random stream gives both sides the same numbers.
Keying each draw by quantities that do not depend on $S$, as Posterior Soup, a composition defined in \cref{sec:supp-operators}, keys its probe noise by the condition, the noise seed, and the step, gives the same numbers as well, and in addition gives the same draw in every subset model, so that images generated with and without a source differ only through the removal; this is a choice of our implementation, not a requirement of exactness.
Keeping the draws made for the full set after a removal, in contrast, reads state from outside the available set and falls outside \cond{3}: if, for example, a composition sampled one source per unit from a single random stream in the order of the sources, the counterfactual model would draw for the remaining sources only, whereas keeping the draws would retain numbers drawn with the removed source in the stream; we do not test such a composition.

\paragraph{Shared configuration and per-source constants.}
The shared configuration $H$ specifies the construction procedure and is the same for every set of sources.
In our experiments, shared constants such as $\kappa$, $w_0$, and the training step are chosen once during development and then fixed as part of $H$.
Every quantity fitted for a single source, including its LoRA, relevance model, and calibration, is computed from its own data, its own component, and source-independent references and stored in $C_g$, so \cond{2} applies to it (\cref{sec:supp-relevance}).
A model built on any subset therefore uses the same procedure and only the components of that subset, and recombination recovers it exactly.
Since the development inputs involve all sources, rerunning this development without the removed sources defines a counterfactual model that also reruns the development.
For the relevance constants and the inpainting threshold, the development is itself a fixed rule applied to the components, so it becomes part of the composition and recombination recovers this counterfactual as well (\cref{sec:supp-recalibration}); the training settings and the pretraining of the base model remain part of $H$.

\paragraph{Recombination and component deletion.}
Recombination reruns the composition rule on $S\setminus B$, so the weights are recomputed and other sources can take over the role of a removed one.
Deleting the terms of $B$ while freezing the weights computed for $S$ answers a narrower question and does not yield the counterfactual model; we therefore use recombination throughout, and \cref{sec:supp-deletion} measures how often the two answers differ.

\paragraph{Other construction choices.}
A quantity trained on the data of all sources falls outside \cond{1} if it is part of the shared model, such as a tuned backbone, outside \cond{2} if it is stored in a component, and outside \cond{3} if the composition reads it during generation, such as a shared selector or pooled statistics; adapters chosen outside the composition rule, such as a hand-picked LoRA fusion, fall outside \cond{3}.
Falling outside the conditions does not by itself show that no exact removal procedure exists; for such constructions, retraining remains exact.

\section{Relevance Scores in Detail}
\label{sec:supp-relevance}

This section details the relevance scores used in our experiments (\cref{sec:relevance}).

\subsection{Text-to-image generation}

\paragraph{Units.}
$\Units(c)$ contains the prompt body (the prompt without its template prefix and final period), each of its words, and each pair of adjacent words, with punctuation other than hyphens and apostrophes removed.
These windows enumerate candidate units without assuming separators or a template, so that a character's name forms a unit regardless of whether the prompt uses commas.

\paragraph{Scores.}
Each unit is embedded as the sentence ``An illustration of $u$.'', the template of the training captions, with the pooled output of SDXL's second text encoder, normalized to unit length.
The relevance score is a kernel density ratio between the embedded captions $\{e^g_i\}_{i=1}^{n_g}$ of source $g$ and a fixed set of $n_{\mathrm{ref}}=154$ short generic prompts (\ifarxivelse{to be released with the code}{included with the code}) $\{e^{\mathrm{ref}}_j\}$ written independently of all sources:
\begin{equation}
  r_g(u)=\frac{\frac{1}{n_g}\sum_{i}\exp\bigl(\kappa\,(e_u^\top e^g_i-1)\bigr)}{\frac{1}{n_{\mathrm{ref}}}\sum_{j}\exp\bigl(\kappa\,(e_u^\top e^{\mathrm{ref}}_j-1)\bigr)}.
  \label{eq:kde}
\end{equation}
The score depends only on the captions of $g$, the text encoder, and the fixed references, as required by \cond{2}; the common reference in the denominator puts the scores of different sources on one scale.
The constants $\kappa$ and $w_0$ are shared by all sources and chosen once on development prompts by a fixed objective that rewards the weights of the covering artist and the named character and penalizes artist weights on prompts without names; they are part of $H$, and \cref{sec:supp-recalibration} reruns this choice on the remaining sources.

\paragraph{Concentration of the weights.}
A larger $\kappa$ sharpens both kernels in \cref{eq:kde}, so that $r_g(u)$ becomes large for sources whose captions are closer to $u$ than the references are and vanishes for the others.
Within a unit, the weights then concentrate on the closest source or on the base model, as with the inverse temperature of a softmax.
The null weight $w_0$ sets how much of each unit falls back to the base model, while the ratio between the weights of two sources on the same unit does not depend on it.
Every pair $(w_0,\kappa)$ shared by all sources defines a fixed construction rule that satisfies the conditions, so varying the constants keeps recombination exact.

\subsection{Inpainting}

The base model is the SDXL inpainting U-Net~\cite{diffusers2023sdxlinpaint}, fine-tuned for inpainting from SDXL~\cite{podell2024sdxl}.
Each source $g$ comes with the part of the cabinet that its data specify, and a region $M_g$ of each training image covers this part: the body from the top of the cabinet to the tops of the legs for the door sources, and the legs for $g_3$, both found from boxes detected by a vision-language model (Qwen3-VL-8B).
The LoRA is trained with
\begin{equation}
  \begin{split}
  \mathcal L_g={}&\operatorname{Mean}_{M_g}\bigl[\|\epsilon_{\theta_0+\Delta\theta_g}-y\|^2\bigr]\\
  &+\lambda_{\mathrm{out}}\operatorname{Mean}_{1-M_g}\bigl[\|\epsilon_{\theta_0+\Delta\theta_g}-\operatorname{sg}(\epsilon_{\theta_0})\|^2\bigr],
  \end{split}
  \label{eq:role-loss}
\end{equation}
where $y$ is the diffusion target, $\operatorname{sg}$ stops the gradient, and $\lambda_{\mathrm{out}}=1$; the second term reads only the frozen base model, so the construction satisfies \cond{2}.
On development images, this loss let a door source and the leg source be used together (16 of 16 images met both specifications, compared with 3 of 16 with the usual loss).
All sources use the same caption, which names no design.

The relevance score uses fixed CLIP ViT-L/14~\cite{radford2021clip} image features of the visible region (the whole image, the top band, and three parts of the band).
For each source, a logistic model on these features is fitted to separate held-out images of $g$ from a fixed set $X_{\mathrm{gen}}$ of 64 generic storage cabinets that names no source attribute, and $r_g$ is its odds.
The LoRA enters \cref{eq:soup} scaled by a strength $a_g$, chosen on held-out images of $g$ alone.
The classifier, its calibration, and $a_g$ are stored in $C_g$, whereas the features, $X_{\mathrm{gen}}$, the regions, the training settings, and $w_0$ are part of $H$.
For a unit $u$, the selected component is $\arg\max_{k}\pi_{k\mid u}$ over the base model, with $\pi_{0\mid u}=w_0/(w_0+\sum_{h}r_h(u))$, and the sources of that part; with $w_0=3/7$, a source is selected when its calibrated probability is at least $0.3$ and higher than those of the other sources of the part.
This threshold and the training step were chosen on development images of all sources, so the guarantee holds with them fixed in $H$; rerunning the choice of the threshold on the remaining sources is exact as well and changes it to $0.5$ for four of the six removals (\cref{sec:supp-recalibration}).

\section{Further Details of the Compositions}
\label{sec:supp-operators}

This section details the compositions of \cref{sec:relevance}; \cref{tab:operators} summarizes them and their cost.

\paragraph{Prior compositions of this form.}
The weights of this form also cover compositions from prior work, built from the same adapters.
Diffusion Soup~\cite{biggs2024soup} uses constant scores and ignores the condition.
The text-to-image variant of CDM~\cite{golatkar2023cdm} scores each source on its predicted clean image, with weights of the form of the first line of \cref{eq:adaptive}; since it sums the noise predictions of the sources instead of merging their parameters, \cref{cor:adaptive} does not apply directly, but it reads only the sources in $S$, so \cond{3} holds and \cref{prop:exact} applies when that image is computed without the removed sources, but it normalizes the weights of all sources over the whole image.
Stylus~\cite{luo2024stylus} selects adapters with a language model; in our adaptation, a fixed language model assigns sources to the keywords of the prompt from descriptions written from each source's own images, which act as constant scores (Stylus-type selection below).

\paragraph{\xsoup.}
Let $u_\ast\in\Units(c)$ be the prompt body and call the other units narrow.
\xsoup keeps the narrow units of \cref{eq:unit-weights} and computes, for each source, its largest weight on a narrow unit, $\nu_g=\max_{u\neq u_\ast}\pi_{g\mid u}$; the sources with $\nu_g\ge1/2$ are explained.
Two sources are substitutes, $g\sim h$, if each is relevant to the captions of the other: with $F_{g\leftarrow h}$ the fraction of the distinct captions $x$ of $h$ with $r_g(x)>w_0$, on which $g$ alone would take more than half of the weight, $g\sim h$ if $\min(F_{g\leftarrow h},F_{h\leftarrow g})\ge\tau$.
The explained-away set $E$ contains the explained sources and every substitute $h$ of an explained source $g$ with $r_h(u_\ast)\le r_g(u_\ast)$, and the prompt body is normalized over the other sources,
\begin{equation}
  \pi_{g\mid u_\ast}=\frac{r_g(u_\ast)\,\mathbf 1[g\notin E]}{w_0+\sum_{h\in S\setminus E}r_h(u_\ast)},
  \label{eq:explainaway}
\end{equation}
before the units are combined by the maximum as in \cref{sec:relevance}.
The name borrows the intuition of explaining away in probabilistic reasoning, that a source already explained by a narrow unit no longer competes for the body, but the rule is not inference in a probabilistic model; the explained sources keep the weights of their narrow units and are still used in generation.
Without the substitutes, a character that the prompt does not name, whose captions resemble those of the named one, would take the share of the body that the named character leaves.
On our sources, $F$ lies between $0.80$ and $1.00$ for two characters, between $0.31$ and $0.44$ for two artists, and between $0.03$ and $0.18$ for an artist and a character, so the substitutes are the same for every $\tau$ between $0.18$ and $0.31$, the characters of each other and the artists of each other; we use $\tau=0.2$.
$F$ reads the relevance of $g$, from $C_g$, on the captions of $h$, from $C_h$, so the weights read only the components of the sources in $S$ and take the form of \cref{eq:adaptive}; after a removal, $F$, the explained sources, and $E$ are computed anew from the remaining sources.
Of the $720$ generations of \cref{tab:t2i} that name a character, the weights differ from those of \soup by more than $0.05$ on $144$; on $38$ of these, some source receives less weight, mostly the second covering artist when the first is explained by a word of the prompt.

\paragraph{Sign-aligned merge.}
Let $R=\{g\in S: w_g\ge0.05\}$ and, for each layer $\ell$ of the LoRAs, $v_{g,\ell}=w_g\,\Delta W_{g,\ell}$ the weighted update of $g\in R$, so that $P_\ell=\sum_{g\in R}v_{g,\ell}$ is the merge of \cref{eq:soup}.
At the steps $i<\lfloor T/2\rfloor$ of the $T=30$ sampling steps, sign-aligned merge replaces $P_\ell$ by
\begin{equation}
  M_\ell=\sum_{g\in R}v_{g,\ell}\odot\mathbf 1\bigl[\operatorname{sign}(v_{g,\ell})=\operatorname{sign}(P_\ell)\bigr],
  \label{eq:signmerge}
\end{equation}
elementwise, which uses the sign election and disjoint selection of TIES-Merging~\cite{yadav2023ties} but sums the kept terms instead of averaging them, applies no scaling factor, and does not trim; the sources outside $R$ enter as in \cref{eq:soup}, and from step $\lfloor T/2\rfloor$ on, the merge is \cref{eq:soup}.
The merge reads only the weights and LoRAs of the sources in $S$ and does not depend on their order, so it is a merge $m_i$ of \cref{eq:adaptive}; with at most one source in $R$, it equals \cref{eq:soup}.
The layers are merged once for each half of the steps, so the network evaluations are those of \soup.
On the $236$ of $240$ generations of D0--D2 ($40$ prompt groups; \cref{sec:supp-exp-t2i}) on which it merges several sources, it drops a median of $17\%$ of the squared norm of the artists' weighted updates and $3\%$ of that of the character.

\paragraph{Posterior Soup.}
Posterior Soup, a composition that we also evaluated (\cref{sec:supp-exp-t2i}), lets the weights depend on the state of generation.
It starts from the weights of \cref{eq:unit-weights}; at every step, the sources that share a unit (competitive units, below) judge the current image by how much better than the base model their LoRAs predict noise added to the current prediction of the clean image, and the weight of the unit shifts toward the sources that explain the image better while its total is kept.
These judgments read only $H$, $c$, $z$, the state, and the component of each source, so the weights take the form of \cref{eq:adaptive}, at the cost of extra network evaluations.
In detail, a unit is competitive if at least two sources in $S$ have $\pi_{g\mid u}>0.01$; only these sources are rescored, and the share of the base model and all other weights stay as in \cref{eq:unit-weights}.
At step $i$, the composition forms $\hat x_0$ from its guided prediction, noises it again to the probe level $t_p=500$ with noise $\eta_i$ obtained by hashing the condition, the noise seed, and $i$, and computes $\ell_0=\|\eta_i-\epsilon_{\theta_0}(x_{t_p},t_p,c)\|^2$ and $\ell_g$ likewise with $\theta_0+\Delta\theta_g$; the improvement is $q^{(i)}_g=(\ell_0-\ell_g)/\ell_0$.
With $R_u$ the rescored sources of a competitive unit $u$, the weights within the unit are reallocated while their total is kept,
\begin{equation}
  \pi^{(i)}_{g\mid u}=\Bigl(\sum_{h\in R_u}\pi_{h\mid u}\Bigr)\frac{\pi_{g\mid u}\,e^{\beta Q^{(i)}_g}}{\sum_{h\in R_u}\pi_{h\mid u}\,e^{\beta Q^{(i)}_h}},\qquad Q^{(i)}_g=\sum_{j\le i}q^{(j)}_g,
  \label{eq:posterior}
\end{equation}
and the units are combined by the maximum as in \cref{sec:relevance}.
We update at every step with $\beta=10$, chosen on development prompts. The accumulated advantage $Q^{(i)}_g$ is computed from $C_g$ and the state, so the weights take the form of \cref{eq:adaptive}.
If no unit is competitive, no probe runs and the image equals that of \soup.
Since $\eta_i$ does not depend on $S$, a removal reruns the same probes on the remaining sources.

\paragraph{Stylus-type selection.}
Our adaptation of Stylus~\cite{luo2024stylus} is not a reproduction of the original pipeline, which retrieves candidates from a large collection of adapters, masks the adapters of each keyword at random, weights them by the refiner's recommended weights, and writes descriptions from model cards; with nine sources, all available sources are presented, so no candidate is lost in retrieval, no mask is applied, and the descriptions are written from the sources' images and captions.
A fixed vision-language model (Qwen3-VL-8B) writes the description of each source from eight of its training images, evenly spaced in the order of their identifiers, and their captions.
The same model, fixed in $H$, reads the prompt and the descriptions of the available sources in the order of their identifiers and assigns zero to five sources to each keyword; following Stylus, keywords that share a source are merged; since we use no recommended weights, each assigned source receives the weight $1/m$, where $m$ is the number of sources assigned to its keyword.
An invalid answer is retried once with a fixed correction prompt and otherwise counted as a failure.
After a removal, the assignment is rerun on the descriptions of the remaining sources, which do not change because each is built from its own source alone.

\paragraph{Storage and time.}
Each text-to-image source stores a LoRA of $93$M parameters ($186$\,MB in bfloat16, $3.6\%$ of the $2.6$B parameters of the SDXL U-Net) together with the caption embeddings of its relevance score; adding a source adds only its component.
On an H100 GPU, a $30$-step generation for a COCO caption takes $1.9$\,s with \soup and $1.8$\,s with the base model.

\begin{table}[t]
  \centering
  \small
  \caption{Compositions under the conditions. Cost counts network evaluations per sampling step relative to the base model with guidance. For Posterior Soup, the cost is the average over the text-to-image evaluation prompts; the Stylus-type selection also calls a vision-language model once per prompt.}
  \label{tab:operators}
  \setlength{\tabcolsep}{4pt}
  \begin{tabular}{@{}lp{0.39\linewidth}c@{}}
    \toprule
    Composition & Composed prediction & Cost \\
    \midrule
    Diffusion Soup & \soup with $w^{(i)}_g(c)=1/|S|$ & $1$ \\
    \soup & merged adapters, \cref{eq:soup} & $1$ \\
    \xsoup & \soup without explained sources in the body, \cref{eq:explainaway} & $1$ \\
    \quad + sign-aligned merge & sign agreement in the first half, \cref{eq:signmerge} & $1$ \\
    Posterior Soup & \soup with weights updated by probes & $1.7$ \\
    Stylus-type & \soup over the assigned sources & $1$ \\
    \bottomrule
  \end{tabular}
\end{table}

\section{Counterfactual Queries}
\label{sec:queries}

Exact recombination turns a question about sources into forward computations on subset models.
We use two queries that answer different questions.

\paragraph{Regeneration.}
Comparing $\Gen_{M_S}(c,z)$ with $\Gen_{M_{S\setminus B}}(c,z)$ for the same prompt and noise reveals whether removing $B$ changes what is drawn, e.g., whether a character disappears, whether another artist's style takes over, or whether the image remains unchanged.

\paragraph{Fixed-image effect.}
For an image with latent $x_0$ and prompt $c$, let
\begin{equation}
  L_S(x_0)=\mathbb{E}_{t,\epsilon}\bigl\|\epsilon-\hat\epsilon_S\bigl(\sqrt{\bar\alpha_t}\,x_0+\sqrt{1-\bar\alpha_t}\,\epsilon,\,t,\,c\bigr)\bigr\|^2
  \label{eq:loss}
\end{equation}
be the denoising loss of the model composed from $S$, whose conditional noise prediction (without guidance) is $\hat\epsilon_S$, estimated with the same draws of $(t,\epsilon)$ for every subset, at the stratified timesteps $t_j=\lfloor 1000\,(j+\frac12)/n\rfloor$, $j=0,\dots,n-1$, for $n$ draws, and with the squared error averaged over the latent elements.
The effect $\Delta_B(x_0)=L_{S\setminus B}(x_0)-L_S(x_0)$ measures how much the model's fit to a fixed image relies on $B$, the quantity targeted by loss-based attribution~\cite{zheng2024dtrak}.
Single removals ($|B|=1$), joint removals, and the difference between them expose substitution and interaction between sources.

\paragraph{Evaluating attribution scores.}
Because recombination gives every subset model exactly, inexpensive attribution scores can be evaluated by their LDS against the losses of the exact subset models on random subsets of the sources without any retraining.
For \cref{sec:exp-attribution}, we use the $552$ prompts of the evaluation set (\cref{sec:supp-exp-t2i}) with the first noise seed, with images generated by the model with all sources, which for nine sources are the images of \cref{tab:t2i}, and $64$ random subsets in which each source is kept with probability $0.5$; every subset loss shares the same $32$ draws of $(t,\epsilon)$.
The attribution scores are the composition weights, D-TRAK~\cite{zheng2024dtrak} with gradients of the composed model, a gradient inner product, and image similarity (CLIP ViT-L/14~\cite{radford2021clip}, DINOv2 ViT-B/14~\cite{oquab2024dinov2}, and CSD~\cite{somepalli2024csd}, averaged over the ten most similar training images), each calibrated on $48$ separate targets.
Throughout, the DINOv2 distance between two images is one minus the cosine similarity of their pooled CLS embeddings from this DINOv2 model.
As references, \cref{tab:lds} also reports the sum of exact single removals, an additive model of the subset losses fitted on one half of the subsets and scored on the other (additive regression), and the agreement between two halves of the loss draws (split-half), which measures the noise of the evaluation.

For these models, approximate attribution scores are not needed, since the effect of removing any set of sources can be computed exactly; the losses of the exact subset models, however, show how such scores behave (\cref{tab:lds}). With \soup, the composition weights predict the subset losses about as well as D-TRAK~\cite{zheng2024dtrak} (LDS $0.70$ and $0.72$ for nine sources, $0.73$ and $0.74$ with the twins), below scoring each source by its exact single removal ($0.78$ and $0.81$) and well above image similarity ($0.40$--$0.50$).
A high LDS shows that a score orders the subset losses well, not that it gives each source its correct share, and we evaluate it only for \soup.
Beyond the aggregate LDS, two observations explain these results and those of \cref{sec:exp-attribution}.
Image similarity assigns most of its positive credit to sources that are unrelated to the prompt.
Redundant sources are invisible to single removals: on the $181$ targets that name a twinned character, removing either twin leaves the character in the regenerated image (identity $0.73$ and $0.76$, compared with $0.75$ with all sources), and only removing both removes it ($0.00$); removing one twin raises the loss of the generated image by $4.7\times10^{-3}$ or $4.6\times10^{-3}$, and removing both by $13\times10^{-3}$.
The composition weights, D-TRAK, and the exact single removals all split the credit for the character evenly between the twins.
The evaluation uses one noise seed and one training run, and the identity of a character is judged automatically.

\begin{table}[t]
  \centering
  \small
  \caption{LDS of attribution scores for \soup against the losses of the exact subset models (mean over the targets with a defined LDS, $520$ to $523$ of the $552$; on the others, all in D4 and D6, which name no character, all subset losses are equal and the LDS is undefined). ``$+2$ twins'' adds $g_{10}$ and $g_{11}$, near-duplicates of the characters $g_5$ and $g_7$, as redundant sources.}
  \label{tab:lds}
  \begin{tabular}{@{}lcc@{}}
    \toprule
    Score & 9 sources & $+2$ twins \\
    \midrule
    Composition weights & 0.70 & 0.73 \\
    D-TRAK~\cite{zheng2024dtrak} & 0.72 & 0.74 \\
    Gradient inner product & 0.66 & 0.70 \\
    CLIP~\cite{radford2021clip} similarity & 0.40 & 0.43 \\
    DINOv2~\cite{oquab2024dinov2} similarity & 0.47 & 0.50 \\
    CSD~\cite{somepalli2024csd} similarity & 0.46 & 0.48 \\
    \midrule
    Exact single removals & 0.78 & 0.81 \\
    Additive regression & 0.77 & 0.81 \\
    Split-half & 0.83 & 0.84 \\
    \bottomrule
  \end{tabular}
\end{table}

\section{Steering Source Dependence by the Composition Rule}
\label{sec:supp-steering}

\paragraph{Rule.}
For a prompt whose subject is covered by two artists $a$ and $b$, the temperature rule keeps, in each unit $u$, the total weight $M_u=\pi_{a\mid u}+\pi_{b\mid u}$ that \cref{eq:unit-weights} gives the two artists, and splits it in proportion to a power of their scores,
\begin{equation}
  \pi^{(\gamma)}_{g\mid u}=M_u\,\frac{r_g(u)^{\gamma}}{r_a(u)^{\gamma}+r_b(u)^{\gamma}},\qquad g\in\{a,b\},
  \label{eq:temperature}
\end{equation}
while the share of the base model and the weights of all other sources stay as in \cref{eq:unit-weights}; a unit in which at most one of the two artists is available and has a positive score keeps the weights of \soup.
The weights are combined over the units by the maximum as in \cref{sec:relevance}.
For $\gamma=0$, the rule splits $M_u$ equally; for $\gamma=1$, it is \soup; and as $\gamma$ grows, it gives $M_u$ to the artist with the higher score.
The two artists are those whose components record the subject of the prompt, and after a removal the rule is computed on the remaining sources, so it reads only the components of the available sources and $H$, and recombination stays exact.
Since the units are combined by the maximum, the total weight of the two artists is kept per unit, not exactly over the whole prompt.
We evaluate $\gamma\in\{0,0.25,0.5,1,2,4,8,16\}$, with the same sources, components, and constants as in \cref{tab:t2i}.
The rule changes only the weights of the two artists; a split over all sources of a unit would, for $\gamma<1$, also move weight onto sources that the unit hardly fits.

\paragraph{Targets and attribution.}
The candidate targets are the $120$ prompts that name a character in a covered subject (D0 of \cref{sec:supp-exp-t2i}), each with two noise seeds, and the $48$ prompts of a covered subject without a name (D5), also with two seeds.
For each target, $a^*$ is the covering artist with the larger weight under \soup and $b^*$ the other, fixed for all $\gamma$.
When the two artists receive the same weight under \soup, up to a relative difference of $10^{-3}$, $a^*$ is undefined; this happens on $2$ of the $240$ targets with a name and on none of the $96$ without, and we report the other $238$ and the $96$ targets.
We fixed this restriction before generating the images of these targets.
For the image $x$ that each rule generates with all sources, we compute the loss of \cref{eq:loss} for every subset of the players $P=\{a^*,b^*,c\}$, with $c$ the named character, or $P=\{a^*,b^*\}$ without a name, while the other sources stay available, with the same $32$ draws of $(t,\epsilon)$ for every subset.
The conditional Shapley value $\phi_g$ is the Shapley value~\cite{shapley1953value,ghorbani2019shapley} of $g\in P$ in the game $v(T)=L_{S\setminus P}(x)-L_{S\setminus(P\setminus T)}(x)$, $T\subseteq P$, which credits the players with the loss reduction $\Delta_P(x)$ that they bring together; it is not the Shapley value over all nine sources.
Intervals are $95\%$ bootstrap intervals over prompts, and differences to $\gamma=1$ are paired per target.

\paragraph{Checks.}
The rule with $\gamma=1$ reproduces the weights of \soup on $600$ prompts, the $552$ of the evaluation set and $48$ that name two characters ($1{,}200$ weight records), up to $4\times10^{-14}$.
For every $\gamma$, on a check target with three removals, the images, weights, and losses equal those of a model built in a separate process from the remaining components, bit for bit.
All targets were completed under every rule.
The answer of the identity judge did not form a valid answer for $5$ of the $1{,}904$ generations of the reported targets, because it repeated its description until the output was cut off; these count as failures.

\paragraph{Results.}
\Cref{tab:steering} and \cref{fig:steering} give the results for the prompts that name a character.
From $\gamma=0$ to $16$, the weight and the credit move from $b^*$ to $a^*$ ($\phi_{a^*}$ from $1.81$ to $4.66$ and $\phi_{b^*}$ from $1.73$ to $-0.29$, $\times10^{-3}$), most of it below $\gamma=1$, and the style share of the covering artists rises from $0.28$ to $0.40$, from $a^*$ ($0.14$ to $0.35$) while that of $b^*$ falls ($0.14$ to $0.05$).
At $\gamma=0$, joint success is $0.155$ $[0.071,0.239]$ lower than at $\gamma=1$ and identity $0.067$ $[0.013,0.122]$ higher; at $\gamma=0.25$ and $0.5$, joint success differs by $-0.076$ $[-0.143,0.000]$ and $-0.034$ $[-0.084,0.017]$; for $\gamma\ge2$, joint success differs by $-0.004$ to $+0.008$ and identity by $-0.013$ to $0.000$, with every interval including zero.
The character keeps its weight, but its value falls from $9.49$ at $\gamma=0$ to $7.79$ at $\gamma=16$ ($-0.31$ $[-0.43,-0.19]$ relative to $\gamma=1$).
The images change by a DINOv2 distance of $0.117$ at $\gamma=0$ and $0.041$--$0.076$ at the other values from those of \soup, less than the distance between two noise seeds on these targets ($0.170$).
Without a name, $\phi_{a^*}$ rises from $1.44$ to $3.13$ and $\phi_{b^*}$ falls from $1.22$ to $-0.01$ ($\times10^{-3}$); at $\gamma=0$, their difference is $0.22$ $[0.06,0.41]$, so the two artists are not credited exactly equally there.

\paragraph{Limits.}
The main text reports averages; for each target, the conditional Shapley values are computed from losses estimated on the exact subset models.
Per target, the credit moves with the weight: on the $382$ steps between consecutive values of $\gamma$ that raise $w_{a^*}$ by more than $0.1$, $\phi_{a^*}$ rises on $99\%$ (median $+1.12\times10^{-3}$), and from $\gamma=0$ to $16$ the difference $\phi_{a^*}-\phi_{b^*}$ grows on $219$ of the $238$ targets, including $158$ of the $160$ on which $w_{a^*}$ rises by at least $0.3$ up to $\gamma=1$.
The steps that change $w_{a^*}$ by less than $0.02$, mostly above $\gamma=1$, change the values by small amounts of either sign (median absolute change $0.04\times10^{-3}$), so $\phi_{a^*}$ never falls over the eight values on only $33$ targets; on the $211$ targets on which it rises from $\gamma=0$ to $16$, its largest drop between consecutive values is in the median $4\%$ of that rise.
At $\gamma=16$, $\phi_{b^*}$ is negative on $158$ targets.
At $\gamma=0$, the two artists receive the same credit only on average; the absolute difference of their values is $0.84$ $[0.74,0.95]$ per target.
The values are signed and are not shares of a fixed total.
Under \soup, the weight of most targets is already concentrated on one artist ($0.76$ vs.\ $0.10$ on average), so temperatures above one change the dependence little.
The temperature cannot separate sources whose scores are equal or nearly equal, such as the twins of \cref{sec:queries}.
The evaluation uses automatic judges, one training run per source, and estimated losses.

\begin{table}[t]
  \centering
  \footnotesize
  \caption{The temperature rule on the $238$ generations that name a character in a subject covered by two artists who receive different weights under \soup; $\gamma=0$ splits their weight equally and $\gamma=1$ is \soup. $\phi$ and $\Delta$ are signed conditional Shapley values and single-removal effects ($\times10^{-3}$); Share: CSD share of the covering artists.}
  \label{tab:steering}
  \setlength{\tabcolsep}{2.1pt}
  \begin{tabular}{@{}cccccccccccc@{}}
    \toprule
    $\gamma$ & $w_{a^*}$ & $w_{b^*}$ & $\phi_{a^*}$ & $\phi_{b^*}$ & $\phi_c$ & $\Delta_{a^*}$ & $\Delta_{b^*}$ & $\Delta_c$ & Joint & ID & Share \\
    \midrule
    0 & 0.41 & 0.41 & 1.81 & 1.73 & 9.49 & 3.72 & 3.45 & 12.29 & 0.43 & 0.86 & 0.28 \\
    0.25 & 0.63 & 0.22 & 3.52 & 0.43 & 8.86 & 6.61 & 1.22 & 12.09 & 0.51 & 0.82 & 0.34 \\
    0.5 & 0.71 & 0.15 & 4.02 & 0.10 & 8.43 & 7.56 & 0.79 & 11.87 & 0.55 & 0.82 & 0.37 \\
    1 & 0.76 & 0.10 & 4.34 & $-0.10$ & 8.10 & 8.18 & 0.53 & 11.66 & 0.58 & 0.79 & 0.38 \\
    2 & 0.79 & 0.07 & 4.54 & $-0.22$ & 7.94 & 8.58 & 0.41 & 11.62 & 0.58 & 0.79 & 0.39 \\
    4 & 0.81 & 0.06 & 4.62 & $-0.27$ & 7.86 & 8.75 & 0.37 & 11.56 & 0.59 & 0.79 & 0.40 \\
    8 & 0.81 & 0.05 & 4.66 & $-0.28$ & 7.81 & 8.84 & 0.35 & 11.55 & 0.58 & 0.78 & 0.40 \\
    16 & 0.82 & 0.05 & 4.66 & $-0.29$ & 7.79 & 8.84 & 0.35 & 11.53 & 0.59 & 0.79 & 0.40 \\
    \bottomrule
  \end{tabular}
\end{table}

\begin{figure*}[t]
  \centering
  \includegraphics[width=\linewidth]{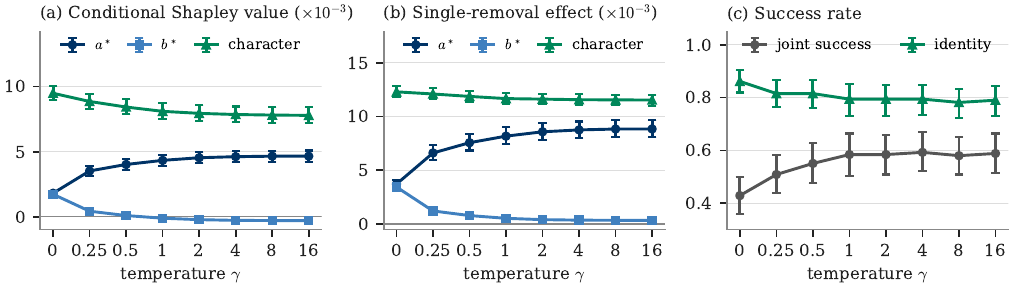}
  \caption{\textbf{Steering source dependence with the temperature rule} (\cref{tab:steering}). Means and $95\%$ bootstrap intervals over the $238$ generations on which the two covering artists receive different weights under \soup; $\gamma=0$ splits their weight equally and $\gamma=1$ is \soup. As $\gamma$ grows, the credit of the artist with the larger weight ($a^*$) rises and that of the other artist ($b^*$) falls from that of $a^*$ to slightly below zero, and the credit of the character declines; joint success is lower and identity higher at $\gamma=0$.}
  \label{fig:steering}
\end{figure*}

\section{Experimental Details}
\label{sec:supp-exp}

\subsection{Text-to-image generation}
\label{sec:supp-exp-t2i}

\paragraph{Sources and prompts.}
The characters $g_5$--$g_9$ are garvo, kettlin, kokebo, momiru, and tessik.
The artists' captions name only the subject, so the style must be learned from the images, whereas the characters' captions name the character; each subject has a list of $24$ scene descriptions, divided into two halves of $12$, one for each of the two artists that cover the subject, so the two artists are trained on captions with different texts.
Each artist covers three of the six covered subjects, and each covered subject is covered by two artists; the artists' images show no people or animals.
All training images are synthetic: SDXL generated the artists' images and reference images of the characters, from which Qwen-Image-Edit generated the character images, and an LLM-based agent drafted part of the evaluation phrases, which were then filtered as described below.
Every source is a LoRA of rank 64 trained for 4{,}000 steps on its own data; images are generated at $1024\times1024$ with $30$ Euler steps and guidance scale $7.0$ in bfloat16. The pooled LoRA has the same rank and is trained for $36{,}000$ steps, nine times as many, on the data of all sources, with the character images repeated so that every source contributes the same number of samples as to its own LoRA.
\ifarxivelse{The implementation and the trained components will be made publicly available.}{A stand-alone implementation of the compositions and of the removal check is included with this supplement; with our components, it reproduces the stored weights to within $6\times10^{-16}$ and the stored images byte for byte (its README), and the trained components will be made publicly available.}
The evaluation prompts, which are never used for training, fall into seven groups: a named character in a covered subject (D0), the same prompt without commas (D1) or paraphrased (D2), a named character in an uncovered subject (D3), an uncovered subject without names (D4), a covered subject without names (D5), and a scene from a character's data without its name (D6).
The covered subjects have 120 groups of D0--D2, 20 for each of the six subjects, and D3--D6 have 48 prompts each, those of D3 and D4 in the two uncovered subjects, vehicles and machines; with two noise seeds, this gives 720 generations in D0--D2 and 1{,}104 in all per method.
The phrases are new and written to be of the kinds of content that the covering artists' data contain; a fixed language model (Phi-4), different from the judge, kept a phrase of a covered subject only if it judged it to be of the same kind as the training data of its subject and to describe no people or animals, and a phrase of an uncovered subject only if it judged it to belong to none of the covered subjects.
We fixed these rules before generating any image of the evaluation set.
CSD measures the share of the covering artists' style by regressing the image embedding on the style centroids of the covering artists, the character, and the base model; an image is a joint success when the judge identifies the character, the share exceeds $0.25$, and the embedding is not closest to the centroid of another source; the last two conditions alone give the style rate of \cref{tab:t2i}.
Intervals in \cref{tab:t2i} and \cref{sec:exp-t2i} are $95\%$ intervals from a stratified cluster bootstrap over prompt groups with $2{,}000$ resamples.
The joint success, identity, and CLIPScore of \cref{tab:t2i} are averaged over D0--D2, and the leakage is measured on D4.
Joint success on the original, comma-free, and paraphrased prompts (D0, D1, D2) is $0.58$, $0.58$, and $0.35$ for \soup, $0.68$, $0.64$, and $0.37$ for \xsoup, $0.70$, $0.69$, and $0.52$ with sign-aligned merge, $0.32$, $0.28$, and $0.19$ for the Stylus-type selection, and $0.25$, $0.26$, and $0.12$ for the pooled LoRA; paraphrasing lowers every method, also in identity, and leaves the order unchanged.
For CLIPScore~\cite{hessel2021clipscore}, we use CLIP ViT-L/14 and $100\max(\cos,0)$, and replace the name of the character in the prompt by ``a character''.
The kernel density score uses $\kappa=160$ and $w_0=30$ (\cref{sec:supp-relevance}).
All judgments in the tables of this experiment are automatic (the Qwen3-VL and Gemma judges, CSD, the LAION improved aesthetic predictor~\cite{schuhmann2022aesthetic,schuhmann2022laion5b} on CLIP ViT-L/14, and DINOv2). Each source is trained once for the results of \cref{tab:t2i} and the other experiments; the comparison of our compositions with the Stylus-type selection and the pooled LoRA is repeated with sources trained from two other seeds on the same data (\cref{sec:exp-t2i}, \cref{sec:supp-baselines}).

\paragraph{Further results.}
Over the generations of \cref{tab:t2i}, \soup exceeds the Stylus-type selection by $+0.24$ $[0.16,0.31]$ in joint success and $+0.19$ $[0.12,0.27]$ in identity, and its leakage differs from that of the base model by $[-0.01,0.02]$.
\xsoup changes joint success by $+0.06$ $[0.02,0.11]$ and identity by $-0.04$ $[-0.08,-0.01]$ relative to \soup; with sign-aligned merge, the changes are $+0.135$ $[0.08,0.19]$ and $+0.08$ $[0.03,0.12]$, at $1.17$ times the time of \soup and a CLIPScore lower by $0.5$ $[0.3,0.7]$.
With sources trained from two other seeds on the same data and prompts, the two exceed \soup in joint success by $+0.09$ $[0.06,0.12]$ and $+0.14$ $[0.10,0.18]$.
On the $115$ prompts of D0 and D2 with the first noise seed on which \soup is a joint success, removing the first covering artist leaves a joint success with the other covering artist on $76$; the character remains on $92$, and the style share of the other artist exceeds $0.25$ on $93$.
\Cref{fig:supp-removal-weights} shows that the weight of a source before a removal does not show the takeover: the weight of the other covering artist is below $0.05$ on $37$ of the $115$ prompts beforehand, yet its CSD share exceeds $0.25$ afterwards on $26$ of them; after the removal its weight is at least $0.9$ on $103$, and its median share rises from $0.16$ to $0.41$ (in the second row of \cref{fig:t2i-qualitative}, from $0.22$ to $0.61$ while its weight stays at $0.97$).

\begin{figure}[t]
  \centering
  \includegraphics[width=\linewidth]{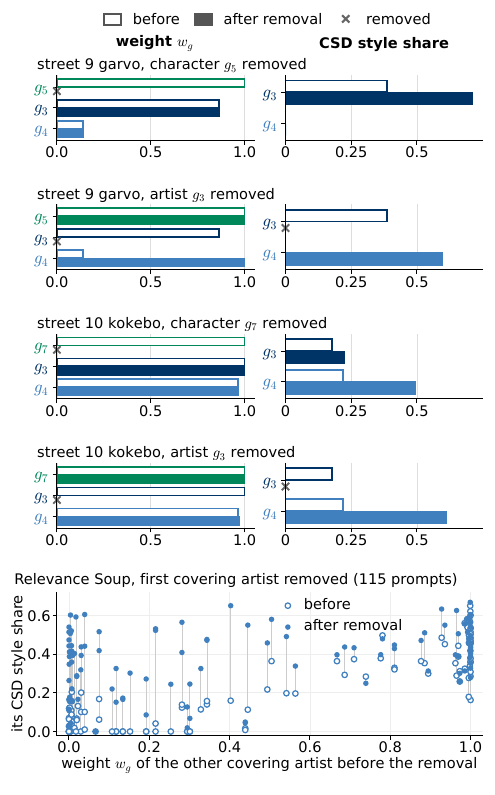}
  \caption{Source weights and style under removal. Top: for the character rows of \cref{fig:t2i-qualitative}, the weight $w_g$ of each source and the CSD style share of the covering artists before (outline) and after (filled) removing the character or the first covering artist from \xsoup with sign-aligned merge. Bottom: on the $115$ prompts of \cref{sec:supp-exp-t2i} on which \soup is a joint success, the CSD share of the other covering artist before and after removing the first one from \soup, against its weight before the removal; the share is computed on the same basis before and after.}
  \label{fig:supp-removal-weights}
\end{figure}

Removing a source unrelated to the prompt changes the images of \soup by a DINOv2 distance of $0.025$ ($0.025$ and $0.016$ for \xsoup without and with sign-aligned merge), compared with $0.151$, $0.067$, and $0.143$ for uniform soup, CDM, and the Stylus-type selection ($48$ prompts of D0 with the first noise seed, each with one artist that does not cover its subject removed; two noise seeds differ by $0.170$ on the targets of \cref{sec:supp-steering}).
The rows of \cref{fig:t2i-qualitative} are chosen by hand among the prompts on which \xsoup with sign-aligned merge is a joint success, and its removals are generated for these rows only; the first character row is one on which \soup fails.
\Cref{tab:t2i-extra} reports measures that \cref{tab:t2i} omits.
Scene is the rate at which the judge, given the scene of the prompt without the name, finds its main setting and objects in the image; every method that draws the character draws the scene less often than the base model, which explains part of its lower CLIPScore.
Tech.\ asks the judge, for each covering artist, whether the background is drawn in that artist's technique; since it accepts a covering artist's technique for $0.32$ of the images of the base model, which uses no artist, above the limit of $0.15$ that we set in advance for the main table, we report it only here.
That limit was set for this judge question; joint success uses the CSD style criterion described above, fixed earlier, which alone accepts $0.29$ of the images of the base model, and also requires the identity, which no image of the base model passes.
The Stylus-type assignment remained invalid for 2 of the 1{,}104 generations, which count as failures; its selector is of the same model family as the Qwen3-VL judge.
On the uncovered subjects without names (D4), it assigns $g_4$ to every generation, alone on $92$ of the $96$, which gives its leakage in \cref{tab:t2i}.
The COCO evaluation uses $5{,}000$ captions of COCO val2017 with one image each; FID is computed with Inception features at $299$ pixels against the center-cropped COCO images, so it is not comparable to FID on $30$k images.
The Stylus-type selection failed on $41$ captions, which are left out of its FID.
The total source weight of \soup is below $0.05$ for $79\%$ of the COCO captions, and its CLIPScore with the COCO caption changes by $-0.13$ $[-0.16,-0.09]$ relative to the base model.
The COCO evaluation thus mainly measures the generations that come close to the base model when no source is relevant; on the prompts that name a character, Scene (\cref{tab:t2i-extra}) and the CLIPScore with the prompt without the name (\cref{tab:t2i}) measure how much of the rest of the prompt is kept.
On the COCO captions, the weights of \xsoup differ from those of \soup by more than $0.05$ on $16$ of the $5{,}000$, and sign-aligned merge changes the merge on $386$; their images equal those of \soup for $99.5\%$ and $92.1\%$ of the captions, and their FIDs are $25.5$ and $25.6$.

\paragraph{Variants of \xsoup.}
\soup, including the rule that sets its constants on development prompts, was fixed before the evaluation prompts were written, so no choice of \soup was made on them, and none lies behind its gains over the other compositions in \cref{tab:t2i} ($+0.24$ $[0.16,0.31]$ over Stylus-type); the explain-away rule and the first-half schedule of sign-aligned merge were chosen with results on these prompts, and on the $480$ generations, two thirds of the prompt groups, not used in the last rounds of that choice, sign-aligned merge exceeds \soup by $+0.13$ $[0.08,0.19]$ and \xsoup by $+0.07$ $[0.02,0.12]$, close to the gains on the other $240$ ($+0.14$ and $+0.08$).
On new prompts built in the same way (D0--D2, two noise seeds, $720$ targets) from $120$ scene phrases that were written and filtered together with those of the evaluation prompts but never used for any generation or choice, with every composition fixed and evaluated once under a protocol fixed in advance, sign-aligned merge exceeds \soup by $+0.15$ $[0.09,0.19]$ in joint success and \xsoup by $+0.05$ $[0.01,0.09]$, \xsoup exceeds \soup by $+0.10$ $[0.05,0.14]$, and \soup exceeds Stylus-type by $+0.23$ $[0.15,0.30]$, all four with Holm-adjusted $p<0.01$; joint success is lower on these prompts by similar amounts for the compositions fixed before the evaluation prompts were written and for those chosen on them (\soup: $0.44$ vs.\ $0.50$; sign-aligned merge: $0.58$ vs.\ $0.64$; Stylus-type: $0.21$ vs.\ $0.26$).
With the sources trained from two other seeds, on the same prompts, the gains hold: pooled over them (\cref{sec:exp-t2i}), the identity of \xsoup does not fall relative to \soup ($-0.00$ $[-0.02,0.01]$), and sign-aligned merge exceeds \xsoup by $+0.05$ $[0.02,0.08]$ in joint success and $+0.06$ $[0.03,0.08]$ in identity.
\Cref{tab:variants} separates the two changes.
Sign alignment at every step on the weights of \xsoup raises identity by $0.10$ $[0.07,0.14]$ but lowers Tech.\ by $0.06$ $[0.03,0.10]$, so joint success rises by only $0.03$ $[-0.01,0.07]$.
In the first half only, sign alignment keeps more of the artists' technique (Tech.\ $+0.03$ $[0.01,0.05]$ over every step) and raises joint success by $0.04$ $[0.01,0.07]$ over sign alignment at every step, and by $0.03$ $[0.01,0.05]$ with the sources of the other two seeds.
In the second half only, it changes joint success by $+0.00$ $[-0.02,0.03]$ and identity by $+0.03$ $[0.01,0.05]$ relative to \xsoup, and its joint success is $0.07$ $[0.03,0.11]$ below that of sign-aligned merge.
On the weights of \soup, sign alignment at every step raises identity to $0.77$ but lowers Tech.\ to $0.57$, and joint success changes by $+0.02$ $[-0.02,0.06]$; in the first half only, it raises joint success by $0.04$ $[0.01,0.08]$ and identity by $0.09$ $[0.05,0.12]$, and by $0.05$ $[0.03,0.08]$ and $0.07$ $[0.05,0.10]$ pooled over the sources of the other two seeds, and \xsoup adds $0.09$ $[0.05,0.13]$ in joint success on top of it ($0.09$ $[0.06,0.12]$ over the other seeds).
Sign alignment in the early steps thus raises identity more than in the late steps, and the weights of \xsoup add most of the gain in joint success.
Posterior Soup, which updates the weights of \soup with the image during generation (\cref{sec:supp-operators}), reaches a joint success of $0.48$ ($-0.03$ $[-0.06,0.00]$ relative to \soup) with identity lower by $0.03$ $[0.01,0.05]$; it costs $1.7$ times the network evaluations and $2.7$ times the time of \soup, and its FID on the COCO captions is $25.6$.

\begin{table}[t]
  \centering
  \small
  \caption{Variants of our compositions on the generations of \cref{tab:t2i} that name a character in a covered subject. Sign alignment is applied at every step, in the first half of the steps (sign-aligned merge, as in \cref{tab:t2i}), or in the second half.}
  \label{tab:variants}
  \setlength{\tabcolsep}{3.2pt}
  \begin{tabular}{@{}lccccc@{}}
    \toprule
    Composition & Joint & ID & Tech. & Share & CLIP \\
    \midrule
    \soup & 0.50 & 0.68 & 0.63 & 0.41 & 24.4 \\
    \quad + sign-aligned, every step & 0.53 & 0.77 & 0.57 & 0.37 & 23.8 \\
    \quad + sign-aligned, first half & 0.55 & 0.77 & 0.59 & 0.38 & 23.9 \\
    \xsoup & 0.57 & 0.64 & 0.65 & 0.45 & 24.5 \\
    \quad + sign-aligned, every step & 0.60 & 0.74 & 0.59 & 0.40 & 23.8 \\
    \quad + sign-aligned, first half & \textbf{0.64} & 0.76 & 0.62 & 0.42 & 23.9 \\
    \quad + sign-aligned, second half & 0.57 & 0.66 & 0.64 & 0.43 & 24.5 \\
    Posterior Soup & 0.48 & 0.65 & 0.59 & 0.41 & 24.5 \\
    \bottomrule
  \end{tabular}
\end{table}

\begin{table}[t]
  \centering
  \small
  \caption{Further text-to-image measures on the generations of \cref{tab:t2i} that name a character in a covered subject (Scene, Tech., Share), on all generations (Aes., the LAION aesthetic score), and on the COCO captions (COCO CLIP). Share is the CSD share of the covering artists' style. Rows as in \cref{tab:t2i}.}
  \label{tab:t2i-extra}
  \setlength{\tabcolsep}{1.8pt}
  \begin{tabular}{@{}lccccc@{}}
    \toprule
    Method & Scene & Tech. & Share & Aes. & COCO\,CLIP \\
    \midrule
    SDXL base & 0.96 & 0.32 & 0.20 & 6.25 & 26.80 \\
    Uniform soup & 0.92 & 0.24 & 0.22 & 6.64 & 27.02 \\
    CDM & 0.93 & 0.82 & 0.75 & 6.15 & 26.11 \\
    Stylus-type & 0.81 & 0.58 & 0.40 & 6.27 & 26.39 \\
    \midrule
    \multicolumn{6}{@{}l}{\emph{Ours}} \\
    \soup & 0.76 & 0.63 & 0.41 & 6.27 & 26.67 \\
    \quad + explain-away & 0.74 & 0.65 & 0.45 & 6.28 & 26.67 \\
    \qquad + sign-aligned merge & 0.73 & 0.62 & 0.42 & 6.27 & 26.67 \\
    \midrule
    Pooled (reference) & 0.81 & 0.43 & 0.32 & 6.15 & 26.74 \\
    \bottomrule
  \end{tabular}
\end{table}

\paragraph{Second judge.}
To check that the comparison does not depend on the judge, we also judge the identity of the generations in \cref{tab:t2i} with Gemma~3 (27B)~\cite{gemmateam2025gemma3}, a vision-language model of another family, keeping CSD and the rest of joint success unchanged.
Gemma is shown a reference image of the character beside the generation, without the list of features given to Qwen3-VL, and asked whether the generation shows the same character design; we chose this question on control images only, without the generations of the compared methods.
Neither judge accepts any of $408$ images of the base model whose prompts name a character or any of $120$ images of a character asked as another character, and on the images of the characters' own data, which the Qwen3-VL judge kept when the data were made, Gemma accepts $216$ of $216$ and Qwen3-VL $215$.
Gemma orders the methods by identity as Qwen3-VL does (\cref{tab:judges}): our three compositions are highest ($0.91$, $0.91$, and $0.94$), followed by the pooled LoRA ($0.80$) and the Stylus-type selection ($0.73$).
Gemma accepts the identity more often than Qwen3-VL: of the $5{,}750$ generations that both judged validly, the two disagree on $868$, and on $864$ of these only Gemma accepts; the human check below finds that Gemma accepts more often than the rater.
With the identity of Gemma in joint success, \xsoup and sign-aligned merge exceed \soup by $+0.106$ $[0.062,0.157]$ and $+0.107$ $[0.060,0.160]$.
Sign-aligned merge raises the Gemma identity of \xsoup by $0.033$ $[0.015,0.053]$ but not this joint success ($+0.001$ $[-0.025,0.028]$), and likewise on the weights of \soup ($+0.040$ $[0.022,0.061]$ and $+0.006$ $[-0.019,0.031]$); Gemma already accepts the identity on $0.91$ of the generations of \xsoup, whereas Qwen3-VL accepts $0.64$.
The gain of sign-aligned merge in joint success thus depends on the identity judge; in the second human sample below, the rater's identity rates for these two compositions are close to those of Qwen3-VL.

\begin{table}[t]
  \centering
  \small
  \caption{Identity rate of the generations of \cref{tab:t2i} as judged by Qwen3-VL (as in \cref{tab:t2i}) and by Gemma~3. Rows as in \cref{tab:t2i}.}
  \label{tab:judges}
  \begin{tabular}{@{}lcc@{}}
    \toprule
    Method & Qwen3-VL & Gemma~3 \\
    \midrule
    SDXL base & 0.00 & 0.00 \\
    Uniform soup & 0.00 & 0.02 \\
    CDM & 0.00 & 0.00 \\
    Stylus-type & 0.48 & 0.73 \\
    \midrule
    \multicolumn{3}{@{}l}{\emph{Ours}} \\
    \soup & 0.68 & 0.91 \\
    \quad + explain-away & 0.64 & 0.91 \\
    \qquad + sign-aligned merge & 0.76 & 0.94 \\
    \midrule
    Pooled (reference) & 0.56 & 0.80 \\
    \bottomrule
  \end{tabular}
\end{table}

\paragraph{Human check of the judges.}
One author judged $300$ outputs, shown in random order without the method, the answers of the judges, or the sampling stratum.
The $200$ text-to-image generations come from D0--D2 of the Stylus-type selection, \soup, Posterior Soup (\cref{sec:supp-operators}), and the pooled LoRA ($2{,}880$ generations), sampled with a fixed seed in three strata by the answers of the two identity judges: $100$ on which they disagree, $50$ that both accept, and $50$ that both reject.
The rater saw reference images of the named character beside the generation and answered whether it shows the same character, with the criterion of the judges.
The $100$ inpainting fills come from the rows of \cref{tab:inpaint}, $50$ that the inpainting judge accepts and $50$ that it rejects, and the rater answered whether a fill meets the specification of its kind, with the criteria of that judge.
The rater never chose the answer ``cannot tell''.
\Cref{tab:human} reports the agreement; for text-to-image, the weighted values reweight the strata by their sizes to estimate the agreement over all $2{,}880$ generations.
The Qwen3-VL judge accepts about as often as the rater answers yes (estimated $0.60$ vs.\ $0.62$), whereas Gemma accepts $0.84$; on the generations where the two judges disagree, almost all accepted by Gemma alone, the rater sides with Qwen3-VL on $64$ of $100$.
Of the fills that the inpainting judge accepts, the rater rejects $11$ of $50$, and of those it rejects, the rater accepts $5$ of $50$; weighting these two strata by their sizes among the $3{,}168$ fills of \cref{tab:inpaint} ($1{,}369$ accepted) gives an estimated agreement of $0.85$ over all fills, close to the $0.84$ of the sample.
A second sample, rated by the same author in the same way, contains $240$ generations of D0--D2 from \xsoup without and with sign-aligned merge ($120$ each; $60$ on which the judges disagree, $30$ that both accept, and $30$ that both reject).
Over these generations, the weighted agreement is $0.82$ (Cohen's $\kappa=0.57$) for Qwen3-VL and $0.77$ ($\kappa=0.32$) for Gemma, and the rater's estimated identity rates, $0.64$ and $0.75$, match those of Qwen3-VL ($0.64$ and $0.76$), so that the rater's estimate of the identity gain of sign-aligned merge is $+0.119$ $[-0.004,0.245]$ (Qwen3-VL: $+0.12$); the rater sides with Qwen3-VL on $72$ of the $120$ generations where the judges disagree and gives the same answer on $9$ of $10$ items repeated from the first sample.
The check covers the identity; the style half of joint success is measured by CSD.
With a single rater, we cannot report agreement between people.

\begin{table}[t]
  \centering
  \small
  \caption{Agreement of the automatic judges with one author on $300$ sampled outputs (accept vs.\ yes). Text-to-image identity: $200$ generations in three strata (judges disagree, both accept, both reject); weighted values estimate the agreement over the $2{,}880$ generations of the four methods that draw the character. Inpainting: $100$ fills, half accepted by the judge.}
  \label{tab:human}
  \setlength{\tabcolsep}{3pt}
  \begin{tabular}{@{}lcccc@{}}
    \toprule
    & \multicolumn{2}{c}{Sample} & \multicolumn{2}{c}{Weighted} \\
    \cmidrule(lr){2-3}\cmidrule(l){4-5}
    Judge & Agree. & $\kappa$ & Agree. & $\kappa$ \\
    \midrule
    Qwen3-VL, identity & 0.80 & 0.55 & 0.85 & 0.69 \\
    Gemma~3, identity & 0.66 & 0.37 & 0.78 & 0.48 \\
    Qwen3-VL, inpainting & 0.84 & 0.68 & -- & -- \\
    \bottomrule
  \end{tabular}
\end{table}

\paragraph{Choices outside the conditions.}
The jointly trained selector is a single multi-class logistic regression on the captions of all sources and generic reference sentences, whose posterior on each unit gives the weights of \soup, as with the classifier of class-conditional CDM; recombination drops the column of the source and renormalizes, whereas the counterfactual model retrains the regression without it.
The weights then differ from those of the counterfactual model in every generation, and the images differ by a DINOv2 distance of $0.056$ when a named source is removed and $0.026$ when an unrelated source is removed, with the identity judgment differing in $4\%$ and $6\%$ of the generations; removing an unrelated source from our composition changes its images by $0.019$ on the $312$ prompts described below.
With the reference statistics of the scores computed from all captions, recombination keeps them while the counterfactual model recomputes them without the removed captions, which changes every \soup weight, and the resulting discrepancy for unrelated sources (DINOv2 distance $0.040$) exceeds the true effect of removing them on the same targets ($0.019$).
These checks use $312$ prompts of the evaluation set, the $120$ of D0 and the $192$ of D3--D6, with one noise seed, removing each of the nine sources in turn.
For the ZipLoRA-type fusion, we optimized the merging coefficients of each pair of a named character and a covering artist on the images of both, with the reconstruction and orthogonality terms of ZipLoRA~\cite{shah2024ziplora} and a fixed number of steps; this adaptation differs from the original in the normalization of the loss and in how activations are collected.
On eight prompts of the evaluation set, removing a source of a fused pair while keeping the coefficients differed from the counterfactual model, which optimizes the coefficients again without the removed source, in all 24 cases, whereas removing an unrelated source matched in all 8.
Computing the coefficients at composition time from statistics that each source stores about its own data would instead satisfy the conditions.
\Cref{sec:supp-deletion} compares the judgments of the selector and the statistics with those of their counterfactual models.

\paragraph{Recombination checks.}
On 24 prompts, each with three removals, recombination was bit-identical to a model instantiated in a separate process from the stored components of the remaining sources only, for \soup and \xsoup, each with and without sign-aligned merge, Posterior Soup, uniform soup, and CDM (72 of 72 each).
The LoRAs of the remaining sources are reused as trained rather than retrained, since by \cond{2} their training does not use the removed sources.
For the Stylus-type selection, on the same $24$ prompts, three runs in separate processes gave identical assignments, the descriptions written again from each source's own images equaled the stored ones, and the assignments and images after a removal equaled those rebuilt from the descriptions of the remaining sources; this holds with greedy decoding, which we set in place of the sampling of the model's default generation configuration.

\paragraph{More sources.}
Adding ten further invented characters, trained like $g_5$--$g_9$, gives $19$ sources; on the targets of \cref{tab:t2i}, no metric of \soup in \cref{tab:scaling} changes by more than $0.018$.
The named character keeps its weight ($0.995$ on average), removing an unrelated source changes the images only slightly more than with nine sources (DINOv2 distance $0.030$ vs.\ $0.025$), and removal still matches rebuilding from the remaining sources bit for bit.
The composition thus takes in more sources without retraining the existing ones or retuning any setting.

\begin{table}[t]
  \centering
  \small
  \caption{\soup with nine versus $19$ sources on the targets of \cref{tab:t2i}; the nine-source row is that of \cref{tab:t2i}. $\Delta$ID is the drop in identity when the named character is removed, over the $816$ generations whose prompts name a character.}
  \label{tab:scaling}
  \begin{tabular}{@{}lccccc@{}}
    \toprule
    Sources & Joint & ID & Leak.$\downarrow$ & Aes. & $\Delta$ID \\
    \midrule
    9  & 0.503 & 0.679 & 0.255 & 6.272 & 0.684 \\
    19 & 0.494 & 0.661 & 0.257 & 6.270 & 0.669 \\
    \bottomrule
  \end{tabular}
\end{table}

\subsection{Inpainting}
\label{sec:supp-exp-inpaint}

\paragraph{Data and evaluation.}
Each source has 58--64 synthetic product photos of cabinets with its design, generated with SDXL, and all sources share a caption that names no design, so a source is defined by its images alone.
The photos of $g_1$ show an arched frame and a diamond medallion in the center of the door, those of $g_2$ a stepped frame and a sunburst medallion, and those of $g_3$ brass caps on the top corners and tall splayed brass legs; their other parts are generic (a plain frame and door, no caps, and short wood or black legs), as in the images that need no source.
Each LoRA is trained only within the region of its part (the cabinet body for the door sources and the legs for $g_3$, \cref{eq:role-loss}), and the relevance of each source is computed from the visible region, where the frame or the caps of its photos appear.
The test images fall into six kinds, which we name by the sets of sources they need: a door design and the legs ($\{g_1,g_3\}$, $\{g_2,g_3\}$), one of them ($\{g_1\}$, $\{g_2\}$, $\{g_3\}$), or neither ($\emptyset$).
The selection rule, its threshold, the training step, and the strengths were frozen on 62 development images.
The test set has 396 images, 150 for each kind that needs two sources and 24 for each other kind, each completed with two noise seeds.
Its images were generated, filtered by a fixed automatic rule, and checked visually with their kinds hidden, by an AI agent rather than a person, before any fill was made; they do not overlap the development images.
A Qwen3-VL judge decides whether a fill shows the door and legs that its kind requires without defects, and a second vision-language model (Claude Sonnet) rates the same criteria from sheets of images shown without the method's name or the kind; under this rater, $91$--$100\%$ of the original images of each kind meet its specification.
Each source is trained once.

\begin{figure*}[t]
  \centering
  \includegraphics[width=\linewidth]{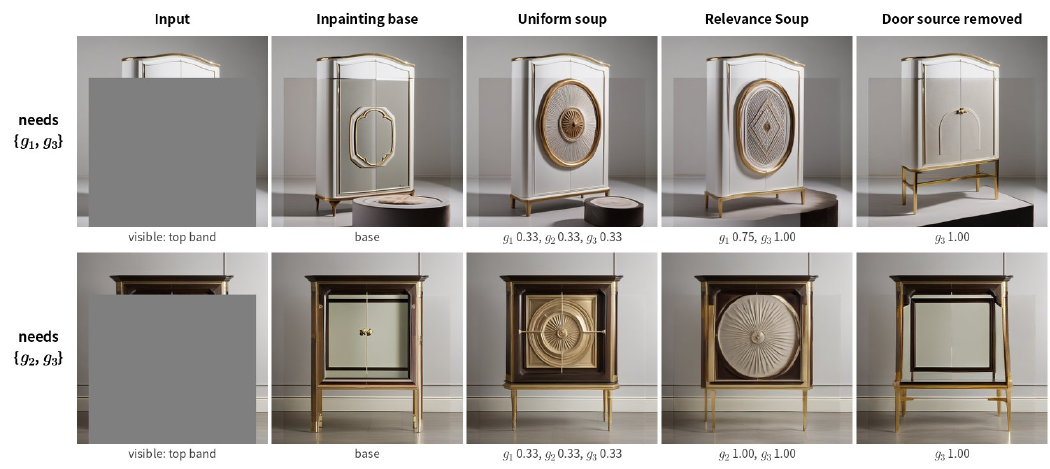}
  \caption{\textbf{Removing a door source in inpainting.}
  The rows of \cref{fig:inpaint-main}, the first test inputs of each kind that needs two sources on which our composition selects both and the judge accepts the fill, with the door source of each kind removed (last column).
  Labels give the sources used and their weights.
  After the removal, the door is left to the base model, while the legs of $g_3$ remain.}
  \label{fig:inpaint-qualitative}
\end{figure*}

\paragraph{Results.}
On the images that need two sources, \soup succeeds on $0.47$, compared with $0.05$ for the base model ($+0.42$ $[0.37,0.47]$), and it exceeds uniform soup by $+0.10$ $[0.05,0.16]$ on these images and $+0.24$ $[0.15,0.33]$ on those that need at most one source.
On the images that need no source, it succeeds on $0.21$, compared with $0.81$ for the base model, because the selection also brings sources into them.
With the second rater, \soup also leads on the images that need at most one source ($0.32$, compared with $0.21$ for the base model, $0.18$ for uniform soup, and $0.62$ for the oracle), so it exceeds uniform soup on both groups with both raters.

\begin{table}[t]
  \centering
  \small
  \caption{Inpainting by kind (\cref{tab:inpaint}). Success under the Qwen3-VL judge / the second rater, averaged over the two seeds of each image; on $\emptyset$ the oracle uses no source and is the base model. Exact: images on which \soup selects exactly the needed sources, out of the images of the kind (the selection does not depend on the seed). Top error: its most common other selection and the number of images (on $\{g_3\}$, $\{g_2,g_3\}$ ties with $\{g_1,g_3\}$).}
  \label{tab:inpaint-kinds}
  \setlength{\tabcolsep}{2.4pt}
  \begin{tabular}{@{}lcccccl@{}}
    \toprule
    & \multicolumn{4}{c}{Success (judge / rater)} & \multicolumn{2}{c}{Selection of ours} \\
    \cmidrule(lr){2-5}\cmidrule(l){6-7}
    Kind & Base & Uniform & Oracle & Ours & Exact & Top error \\
    \midrule
    $\{g_1,g_3\}$ & .05/.01 & .15/.15 & .70/.55 & .30/.25 & 61/150 & $\{g_2,g_3\}$ 33 \\
    $\{g_2,g_3\}$ & .04/.01 & .57/.34 & .88/.71 & .64/.52 & 100/150 & $\{g_2\}$ 33 \\
    $\{g_1\}$     & .04/.00 & .13/.15 & .77/.50 & .44/.27 & 13/24 & $\{g_2\}$ 6 \\
    $\{g_2\}$     & .08/.06 & .65/.52 & .85/.88 & .63/.56 & 15/24 & $\{g_2,g_3\}$ 5 \\
    $\{g_3\}$     & .75/.25 & .04/.06 & .94/.58 & .52/.33 & 11/24 & $\{g_1,g_3\}$ 6 \\
    $\emptyset$   & .81/.52 & .02/.00 & .81/.52 & .21/.10 & 4/24 & $\{g_1\}$ 8 \\
    \bottomrule
  \end{tabular}
\end{table}

\paragraph{By kind.}
\Cref{tab:inpaint-kinds} breaks \cref{tab:inpaint} down by kind: on each kind that needs a source, \soup selects exactly the needed sources on $41$--$67\%$ of the images, and its most common error is the wrong door source on $\{g_1,g_3\}$ and a missing $g_3$ on $\{g_2,g_3\}$.
When the selection is exact, the weights are those of the oracle and the fills are identical in pixels ($408$ of the $792$ fills), so the gap to the oracle comes entirely from the other images, on which the judge accepts only $0.05$--$0.19$ of the fills.
On the images that need no source, it selects none on only $4$ of $24$ and brings in $g_1$ or $g_3$ on the other $20$, so it succeeds on $0.21$ with the judge and $0.10$ with the second rater, compared with $0.81$ and $0.52$ for the base model.
Errors in the selection of sources thus account for the gap to the oracle, which uses the same composition; closing it requires recognizing the needed parts from the visible region better, and a stronger relevance score, fitted like ours to each source's own images against fixed generic ones, would stay within the conditions.

\paragraph{Removal.}
On the test set, recombination after removing $g_1$ or $g_3$ matched a model built in a separate process in pixels and weights on 8 inputs each for our composition and uniform soup (32 of 32), where uniform soup gives each remaining source the same weight.
Of the $792$ fills of \soup over all kinds and both seeds, $189$ show the medallion and $422$ the sunburst, including fills of kinds that do not need it; removing $g_1$ removes the medallion from $183$ of these $189$, and removing $g_2$ removes the sunburst from $399$ of these $422$ (\cref{fig:inpaint-qualitative}).

\section{Source Removal with Recalibration}
\label{sec:supp-recalibration}

\paragraph{Recalibration as part of the composition.}
A calibration rule $Q$ reads the components $\{(g,C_g)\}_{g\in S}$ of the available sources and development inputs that do not depend on the sources, and returns shared constants $h=Q(S)$.
The composition with recalibration,
\begin{equation}
  \Phi'_H\bigl(\{(g,C_g)\}_{g\in S};\,c,z\bigr)=\Phi_{H,\,Q(S)}\bigl(\{(g,C_g)\}_{g\in S};\,c,z\bigr),
\end{equation}
depends only on the components of $S$ and on $H$, so it satisfies \cond{3}, and \cref{prop:exact} applies: recombination equals the construction that reruns $Q$ on $S\setminus B$.
For text-to-image, the development prompts combine the subjects covered by the remaining artists, which each artist's component records, with generic subjects that no artist covers, and they name only the remaining characters.
For inpainting, $Q$ uses the development images of the kinds whose needed sources all remain.

\paragraph{Procedure.}
Each rule was frozen before it was run and repeats how the constants of the paper were chosen.
For text-to-image, $Q$ selects $(\kappa,w_0)$ from $\kappa\in\{10,20,40,80,160\}$ and $w_0\in\{0.1,0.3,1,3,10,30,100\}$ by the development objective of \cref{sec:supp-relevance}.
For inpainting, it selects the threshold of our composition from $\{0.3,0.4,0.5\}$ by the success rate on the development images, averaged over kinds.
We rerun $Q$ after $17$ removals in text-to-image (each of the nine sources, the six pairs of artists that cover a subject together, all artists, and all characters) and after six in inpainting (each source and each pair); no adapter is retrained.

\paragraph{Results.}
On all sources, both rules return the constants of the paper.
In text-to-image, the constants stay at $(\kappa,w_0)=(160,30)$ for $14$ of the $17$ removals, including all characters and five of the six pairs of artists, so the two counterfactuals coincide there; removing $g_9$, the pair $\{g_2,g_3\}$, or all artists changes $w_0$ to $10$, with small effects on the generated images (\cref{tab:recalibration}).
For $\{g_2,g_3\}$, the choice is close: the development objective at $w_0=10$ exceeds that at $w_0=30$ by $0.0004$.
In inpainting, the threshold changes from $0.3$ to $0.5$ for four of the six removals.
The success averaged over kinds changes by $+0.014$ to $+0.034$, but a higher threshold brings in fewer sources, so single kinds move in opposite directions: after removing $g_1$, the success falls by $0.15$ on kind $\{g_2,g_3\}$ and rises by $0.15$ and $0.10$ on kinds $\{g_3\}$ and $\emptyset$.
Recombination gives both counterfactuals exactly, and they differ only where the constants change.
Removal with recalibration matched a rebuild in a separate process that loads only the remaining sources: the selected constants for every set, the weights, and the images ($6$ text-to-image pairs and $17$ inpainting fills) were identical.
Rerunning $Q$ takes about $5$\,s of CPU time per set in text-to-image and at most $32$ new development fills per set in inpainting.
The training settings, the features, and the reference data stay in $H$, and each remaining source keeps its strength $a_g$ in $C_g$.
The small inpainting development set ($15$--$39$ images after a removal) makes the selected threshold sensitive to a few images.

\begin{table}[t]
  \centering
  \footnotesize
  \caption{Constants $Q(S\setminus B)$ after removal and the effect of recalibration (R) compared with the fixed constants (F) on the evaluation sets, with $95\%$ intervals. Text-to-image: $(\kappa,w_0)$, with the mean DINOv2 distance between R and F over $112$ evaluation prompts ($120$ for $\{g_2,g_3\}$); with development prompts built from all eight subjects instead, $w_0$ changes after removing $g_9$ or all artists (to $10$ and $100$) but not after removing $\{g_2,g_3\}$. Inpainting: the threshold, with the inputs whose weights change (of $396$) and the change in success (All, Two, and One/none as in \cref{tab:inpaint}).}
  \label{tab:recalibration}
  \setlength{\tabcolsep}{2.5pt}
  \begin{tabular}{@{}llll@{}}
    \toprule
    Removed & $Q(S\setminus B)$ & Changed & R $-$ F \\
    \midrule
    \multicolumn{4}{@{}l}{\emph{Text-to-image}} \\
    none, 14 sets & $(160,30)$ & -- & identical \\
    $g_9$ & $(160,10)$ & $0.032$ & ID $-0.031$ $[-0.125,0.047]$ \\
    $\{g_2,g_3\}$ & $(160,10)$ & $0.029$ & ID $+0.033$ $[-0.025,0.108]$ \\
    $g_1$--$g_4$ & $(160,10)$ & $0.023$ & ID $-0.047$ $[-0.115,0.000]$ \\
    \midrule
    \multicolumn{4}{@{}l}{\emph{Inpainting}} \\
    none, $g_2$, & $0.3$ & -- & identical \\
    $\{g_2,g_3\}$ & & & \\
    $g_1$ & $0.5$ & $94$ & All $+0.016$ $[-0.023,0.058]$ \\
    & & & Two $-0.067$ $[-0.095,-0.042]$ \\
    $g_3$ & $0.5$ & $64$ & All $+0.034$ $[-0.009,0.074]$ \\
    & & & One/none $+0.068$ $[0.005,0.130]$ \\
    $\{g_1,g_2\}$ & $0.5$ & $48$ & All $+0.017$ $[0.000,0.038]$ \\
    $\{g_1,g_3\}$ & $0.5$ & $54$ & All $+0.014$ $[-0.016,0.046]$ \\
    \bottomrule
  \end{tabular}
\end{table}

\section{Recombination versus Component Deletion}
\label{sec:supp-deletion}

Component deletion (\cref{sec:supp-proof}) removes the terms of $B$ and keeps the weights computed for $S$; like recombination, it costs one generation.
This section measures how often its answer differs from that of the counterfactual model, which recombination gives exactly, in the judgments of \cref{tab:t2i}.

\begin{table}[t]
  \centering
  \footnotesize
  \caption{Component deletion (Del.) versus recombination (Rec.), the counterfactual model, on the same targets: the rate of the judgment after the removal and the share of targets on which the two images are judged differently (Diff., $95\%$ interval), for \soup and for \xsoup with sign-aligned merge ($^\ast$). Joint: joint success; ID: identity. For (c), joint success is defined on the $720$ removals in D0; identity differs on $2\%$ of the $1{,}104$ removals from prompts that name a character for both compositions. Removals as in \cref{sec:supp-deletion}.}
  \label{tab:deletion}
  \setlength{\tabcolsep}{2pt}
  \begin{tabular}{@{}llrcccc@{}}
    \toprule
    Removed & & $n$ & Rec. & Del. & Diff. & Diff.$^\ast$ \\
    \midrule
    (a) a covering artist & Joint & $115$ & 0.66 & 0.37 & 0.36 {\scriptsize[.27,.44]} & 0.45 {\scriptsize[.37,.54]} \\
    (b) named character & ID & $120$ & 0.00 & 0.00 & 0.00 & 0.00 \\
    (c) an unrelated source & Joint & $720$ & 0.61 & 0.61 & 0.02 {\scriptsize[.01,.04]} & 0.03 {\scriptsize[.01,.04]} \\
    (d) one twin & ID & $362$ & 0.75 & 0.03 & 0.72 {\scriptsize[.67,.77]} & -- \\
    (d) both twins & ID & $181$ & 0.00 & 0.00 & 0.00 & -- \\
    (e) both covering artists & ID & $115$ & 0.93 & 0.92 & 0.06 {\scriptsize[.03,.10]} & 0.10 {\scriptsize[.05,.17]} \\
    \bottomrule
  \end{tabular}
\end{table}

\paragraph{Setup.}
For each target, both procedures start from the same prompt and noise, and we compare the identity and joint success of the two images (\cref{sec:supp-exp-t2i}).
The removals are (a) the first covering artist, on the $115$ prompts of \cref{sec:supp-exp-t2i} on which \soup is a joint success; (b) the named character, on the $120$ prompts of D0; (c) each source other than the named character and the covering artists, on the $312$ prompts of \cref{sec:supp-exp-t2i} ($2{,}304$ removals); (d) one twin or both, on the $181$ targets of \cref{sec:queries} that name a twinned character; and (e) both covering artists, on the prompts of (a).
All targets use the first noise seed; the twins (d) are added to \soup only, and for \xsoup with sign-aligned merge, (a) and (e) use the same $115$ prompts.

\paragraph{Results.}
\Cref{tab:deletion} shows that deletion and recombination agree where no other source can take the role of the removed one: after removing the named character (b) or both twins (d), neither keeps the character, and after removing an unrelated source (c), the two images are judged differently in $2\%$ of the removals.
They disagree where another source can take over.
After removing a covering artist (a), the other covering artist takes over in the counterfactual model on $0.66$ of the prompts but only on $0.37$ under deletion, which keeps the weight that this artist had beside the removed one, so joint success differs on $36\%$ of the prompts ($45\%$ for \xsoup with sign-aligned merge).
After removing one twin (d), the counterfactual model keeps the character on $0.75$ of the targets, whereas deletion keeps it on $0.03$; the character survives either single removal on $59\%$ of the targets in the counterfactual model and on $1\%$ under deletion, so deletion would report each twin as necessary.
Which case applies to a generation is not known before the counterfactual is computed.
The fixed-image effects of \cref{sec:queries} computed with deletion change which of the two covering artists and the character has the largest conditional Shapley value (\cref{sec:supp-steering}) on $13\%$ $[10\%,17\%]$ of the $357$ targets of D0--D2 that name a character and whose two artists receive different weights (first noise seed).

\paragraph{Choices outside the conditions.}
The same comparison applies to the two choices outside the conditions of \cref{sec:supp-exp-t2i}, on their $312$ prompts, now between recombination and their counterfactual models.
With the jointly trained selector kept after a removal, whether the removal changes the identity or style judgment differs from the counterfactual model on $8.0\%$ $[5.4\%,11.0\%]$ of the removals of a covering artist and $6.2\%$ $[4.8\%,7.8\%]$ of the removals of an unrelated source; with the reference statistics kept, it differs on $9.8\%$ $[7.1\%,12.8\%]$ and $3.5\%$ $[2.0\%,5.2\%]$.
For unrelated sources, the counterfactual removal itself changes these judgments on $7.7\%$ and $4.1\%$ of the same removals, so the error is about as large as the effect it would measure.

\section{Further Baselines and Ablations}
\label{sec:supp-baselines}

\Cref{tab:baselines} compares further compositions on the generations of \cref{tab:t2i}, with the same judges and sources trained once.
Each composition and its constants were fixed before its images were generated, and for each except the hand-given sources, which are references, recombination matched a model built in a separate process (72 of 72, as in \cref{sec:supp-exp-t2i}).

\begin{table}[t]
  \centering
  \footnotesize
  \caption{Further compositions on the generations of \cref{tab:t2i}; columns as there. Given sources: the hand-given references, defined only where a character is named; word units: words and pairs of adjacent words. Our compositions and uniform soup at $1/9$ repeat \cref{tab:t2i}.}
  \label{tab:baselines}
  \setlength{\tabcolsep}{1.2pt}
  \begin{tabular}{@{}lcccccc@{}}
    \toprule
    Method & Joint$\uparrow$ & ID$\uparrow$ & Style$\uparrow$ & CLIP & Leak.$\downarrow$ & FID$\downarrow$ \\
    \midrule
    \soup & 0.50 & 0.68 & 0.78 & 24.4 & 0.26 & 25.5 \\
    \enspace + explain-away & 0.57 & 0.64 & 0.88 & 24.5 & 0.26 & 25.5 \\
    \quad\enspace + sign-aligned merge & 0.64 & 0.76 & 0.85 & 23.9 & 0.26 & 25.6 \\
    \midrule
    Given sources, each $1$ & 0.30 & 0.30 & 0.99 & 24.8 & -- & -- \\
    \enspace artists share $1$ & 0.45 & 0.75 & 0.66 & 24.7 & -- & -- \\
    \midrule
    Uniform soup, each $1/9$ & 0.00 & 0.00 & 0.34 & 30.3 & 0.22 & 31.5 \\
    \enspace each $0.25$ & 0.00 & 0.00 & 0.36 & 26.5 & 0.21 & 48.2 \\
    \enspace each $0.5$ & 0.01 & 0.01 & 0.52 & 24.9 & 0.46 & 132.7 \\
    \enspace each $1$ & 0.00 & 0.00 & 0.39 & 16.6 & 1.00 & 285.8 \\
    \midrule
    Retrieval-type, $\tau_{\mathrm{r}}=0.65$ & 0.01 & 0.10 & 0.40 & 28.4 & 0.25 & 24.7 \\
    \enspace without threshold & 0.25 & 0.26 & 0.89 & 24.8 & 0.54 & 81.1 \\
    \midrule
    \soup, prompt unit only & 0.00 & 0.09 & 0.86 & 26.4 & 0.26 & 25.9 \\
    \soup, word units only & 0.20 & 0.79 & 0.30 & 24.2 & 0.24 & 25.3 \\
    \bottomrule
  \end{tabular}
\end{table}

\paragraph{Compositions.}
The hand-given sources use the sources that the prompt needs, given by hand rather than selected, so they are defined only on prompts that name a character in a covered subject: the named character and each covering artist at weight $1$, or the character at weight $1$ and the two covering artists sharing a weight of $1$.
Uniform soup scales every LoRA by $0.25$, $0.5$, or $1$ instead of the $1/9$ of \cref{tab:t2i}.
The retrieval-type composition follows LoraRetriever~\cite{zhao2024loraretriever} without its tuned retriever: each source stores the normalized mean of the embeddings of its captions, with the encoder and template of \cref{sec:supp-relevance}, and the $k=3$ sources closest to the prompt, among those with cosine above $\tau_{\mathrm{r}}$, are merged at weight $1$ rather than averaged as in LoraRetriever; $k$ and $\tau_{\mathrm{r}}=0.65$ were chosen on the development prompts of \cref{sec:supp-relevance} by the same objective as $(\kappa,w_0)$, and we also report it without the threshold.
The two unit ablations restrict \soup to the whole prompt or to words and pairs of adjacent words, with $(\kappa,w_0)$ chosen again by the same objective ($(160,3)$ and $(80,0.3)$).

\paragraph{Results.}
The hand-given sources do not reach the joint success of our compositions.
At weight $1$ each, the artists override the character (identity $0.30$, style $0.99$); with the artists sharing a weight, joint success is $0.45$, on par with \soup ($-0.05$ $[-0.13,0.03]$) and below \xsoup with sign-aligned merge ($-0.18$ $[-0.26,-0.12]$), with identity similar to the latter ($-0.01$ $[-0.05,0.04]$) and a lower style rate.
None of the four tested scales lets uniform soup draw the character: joint success stays at most $0.01$, while FID rises to $286$ when every LoRA is at weight $1$.
With its threshold, the retrieval-type composition falls back to the base model on $71\%$ of the prompts that name a character and on all prompts without a fitting source, including every COCO caption, because the development prompts follow a template that the evaluation prompts do not; without the threshold, it merges three sources at weight $1$ for every prompt and reaches a joint success of $0.25$ with a leakage of $0.54$ and an FID of $81$.
The whole prompt alone finds the artist but rarely the character (identity $0.09$), and words and pairs alone find the character but not the artist (style $0.30$, near the base model), so \soup needs both kinds of unit.
Among the constants, halving or doubling $w_0$ changes the joint success of \soup by at most $0.01$, whereas halving $\kappa$ lowers it by $0.15$ $[0.09,0.21]$; $\kappa=160$ is the largest value of the grid of \cref{sec:supp-recalibration} and is selected after every removal there.
With sources trained from three seeds, joint success is $0.52\pm0.02$ for \soup, $0.26\pm0.02$ for the Stylus-type selection, and $0.22\pm0.03$ for the pooled LoRA of \cref{tab:t2i} (mean $\pm$ standard deviation over the seeds); over $18$ pooled LoRAs, from three seeds, three training lengths ($12$k, $24$k, and $36$k steps), and captions with or without identifiers of the sources, which are then added to the prompt for the needed sources, it ranges from $0.15$ to $0.36$, below \soup with the sources of every seed.

%% file: main.bbl
\begin{thebibliography}{74}
\providecommand{\natexlab}[1]{#1}
\providecommand{\url}[1]{\texttt{#1}}
\expandafter\ifx\csname urlstyle\endcsname\relax
  \providecommand{\doi}[1]{doi: #1}\else
  \providecommand{\doi}{doi: \begingroup \urlstyle{rm}\Url}\fi

\bibitem[Alberti et~al.(2025)Alberti, Hasanaliyev, Shah, and
  Ermon]{alberti2025siss}
Silas Alberti, Kenan Hasanaliyev, Manav Shah, and Stefano Ermon.
\newblock Data unlearning in diffusion models.
\newblock In \emph{International Conference on Learning Representations
  (ICLR)}, 2025.

\bibitem[Bai et~al.(2025)Bai, Cai, Chen, Chen, Chen, et~al.]{bai2025qwen3vl}
Shuai Bai, Yuxuan Cai, Ruizhe Chen, Keqin Chen, Xionghui Chen, et~al.
\newblock {Qwen3-VL} technical report.
\newblock \emph{arXiv preprint arXiv:2511.21631}, 2025.

\bibitem[Biggs et~al.(2024)Biggs, Seshadri, Zou, Jain, Golatkar, Xie, Achille,
  Swaminathan, and Soatto]{biggs2024soup}
Benjamin Biggs, Arjun Seshadri, Yang Zou, Achin Jain, Aditya Golatkar, Yusheng
  Xie, Alessandro Achille, Ashwin Swaminathan, and Stefano Soatto.
\newblock {Diffusion Soup}: Model merging for text-to-image diffusion models.
\newblock In \emph{European Conference on Computer Vision (ECCV)}, 2024.

\bibitem[Bourtoule et~al.(2021)Bourtoule, Chandrasekaran, Choquette-Choo, Jia,
  Travers, Zhang, Lie, and Papernot]{bourtoule2021unlearning}
Lucas Bourtoule, Varun Chandrasekaran, Christopher~A. Choquette-Choo, Hengrui
  Jia, Adelin Travers, Baiwu Zhang, David Lie, and Nicolas Papernot.
\newblock Machine unlearning.
\newblock In \emph{IEEE Symposium on Security and Privacy (S\&P)}, 2021.

\bibitem[Bowman et~al.(2023)Bowman, Achille, Zancato, Trager, Perera, Paolini,
  and Soatto]{bowman2023alacarte}
Benjamin Bowman, Alessandro Achille, Luca Zancato, Matthew Trager, Pramuditha
  Perera, Giovanni Paolini, and Stefano Soatto.
\newblock {{\`A}-la-carte} prompt tuning ({APT}): Combining distinct data via
  composable prompting.
\newblock In \emph{IEEE/CVF Conference on Computer Vision and Pattern
  Recognition (CVPR)}, 2023.

\bibitem[Chowdhury et~al.(2025)Chowdhury, Choromanski, Sehanobish, Dubey, and
  Chaturvedi]{chowdhury2025s3t}
Somnath Basu~Roy Chowdhury, Krzysztof Choromanski, Arijit Sehanobish, Avinava
  Dubey, and Snigdha Chaturvedi.
\newblock Towards scalable exact machine unlearning using parameter-efficient
  fine-tuning.
\newblock In \emph{International Conference on Learning Representations
  (ICLR)}, 2025.

\bibitem[Dai and Gifford(2026)]{dai2026unattributable}
Zheng Dai and David~K. Gifford.
\newblock Outputs of generative diffusion models are often unattributable.
\newblock \emph{Nature Communications}, 17:\penalty0 6974, 2026.

\bibitem[Du et~al.(2023)Du, Durkan, Strudel, Tenenbaum, Dieleman, Fergus,
  Sohl-Dickstein, Doucet, and Grathwohl]{du2023reduce}
Yilun Du, Conor Durkan, Robin Strudel, Joshua~B. Tenenbaum, Sander Dieleman,
  Rob Fergus, Jascha Sohl-Dickstein, Arnaud Doucet, and Will Grathwohl.
\newblock Reduce, reuse, recycle: Compositional generation with energy-based
  diffusion models and {MCMC}.
\newblock In \emph{International Conference on Machine Learning (ICML)}, 2023.

\bibitem[Dukler et~al.(2023)Dukler, Bowman, Achille, Golatkar, Swaminathan, and
  Soatto]{dukler2023safe}
Yonatan Dukler, Benjamin Bowman, Alessandro Achille, Aditya Golatkar, Ashwin
  Swaminathan, and Stefano Soatto.
\newblock {SAFE}: Machine unlearning with shard graphs.
\newblock In \emph{IEEE/CVF International Conference on Computer Vision
  (ICCV)}, 2023.

\bibitem[Fan et~al.(2024)Fan, Liu, Zhang, Wong, Wei, and Liu]{fan2024salun}
Chongyu Fan, Jiancheng Liu, Yihua Zhang, Eric Wong, Dennis Wei, and Sijia Liu.
\newblock {SalUn}: Empowering machine unlearning via gradient-based weight
  saliency in both image classification and generation.
\newblock In \emph{International Conference on Learning Representations
  (ICLR)}, 2024.

\bibitem[Gandikota et~al.(2023)Gandikota, Materzynska, Fiotto-Kaufman, and
  Bau]{gandikota2023erasing}
Rohit Gandikota, Joanna Materzynska, Jaden Fiotto-Kaufman, and David Bau.
\newblock Erasing concepts from diffusion models.
\newblock In \emph{IEEE/CVF International Conference on Computer Vision
  (ICCV)}, 2023.

\bibitem[{Gemma Team}(2025)]{gemmateam2025gemma3}
{Gemma Team}.
\newblock Gemma 3 technical report.
\newblock \emph{arXiv preprint arXiv:2503.19786}, 2025.

\bibitem[Georgiev et~al.(2023)Georgiev, Vendrow, Salman, Park, and
  Madry]{georgiev2023journey}
Kristian Georgiev, Joshua Vendrow, Hadi Salman, Sung~Min Park, and Aleksander
  Madry.
\newblock The journey, not the destination: How data guides diffusion models.
\newblock \emph{arXiv preprint arXiv:2312.06205}, 2023.

\bibitem[Ghorbani and Zou(2019)]{ghorbani2019shapley}
Amirata Ghorbani and James Zou.
\newblock Data {Shapley}: Equitable valuation of data for machine learning.
\newblock In \emph{International Conference on Machine Learning (ICML)}, 2019.

\bibitem[Ghosal et~al.(2026)Ghosal, Maini, and Raghunathan]{ghosal2026nulls}
Gaurav~R. Ghosal, Pratyush Maini, and Aditi Raghunathan.
\newblock Natively unlearnable large language models.
\newblock \emph{arXiv preprint arXiv:2606.13873}, 2026.

\bibitem[Golatkar et~al.(2023)Golatkar, Achille, Swaminathan, and
  Soatto]{golatkar2023cdm}
Aditya Golatkar, Alessandro Achille, Ashwin Swaminathan, and Stefano Soatto.
\newblock Training data protection with compositional diffusion models.
\newblock \emph{arXiv preprint arXiv:2308.01937}, 2023.

\bibitem[Gu et~al.(2023)Gu, Wang, Wu, Shi, Chen, Fan, Xiao, Zhao, Chang, Wu,
  et~al.]{gu2023mixofshow}
Yuchao Gu, Xintao Wang, Jay~Zhangjie Wu, Yujun Shi, Yunpeng Chen, Zihan Fan,
  Wuyou Xiao, Rui Zhao, Shuning Chang, Weijia Wu, et~al.
\newblock {Mix-of-Show}: Decentralized low-rank adaptation for multi-concept
  customization of diffusion models.
\newblock In \emph{Advances in Neural Information Processing Systems
  (NeurIPS)}, 2023.

\bibitem[Hessel et~al.(2021)Hessel, Holtzman, Forbes, Le~Bras, and
  Choi]{hessel2021clipscore}
Jack Hessel, Ari Holtzman, Maxwell Forbes, Ronan Le~Bras, and Yejin Choi.
\newblock {CLIPScore}: A reference-free evaluation metric for image captioning.
\newblock In \emph{Conference on Empirical Methods in Natural Language
  Processing (EMNLP)}, 2021.

\bibitem[Heusel et~al.(2017)Heusel, Ramsauer, Unterthiner, Nessler, and
  Hochreiter]{heusel2017fid}
Martin Heusel, Hubert Ramsauer, Thomas Unterthiner, Bernhard Nessler, and Sepp
  Hochreiter.
\newblock {GANs} trained by a two time-scale update rule converge to a local
  {Nash} equilibrium.
\newblock In \emph{Advances in Neural Information Processing Systems
  (NeurIPS)}, 2017.

\bibitem[Ho and Salimans(2022)]{ho2022cfg}
Jonathan Ho and Tim Salimans.
\newblock Classifier-free diffusion guidance.
\newblock \emph{arXiv preprint arXiv:2207.12598}, 2022.

\bibitem[Hu et~al.(2022)Hu, Shen, Wallis, Allen-Zhu, Li, Wang, Wang, and
  Chen]{hu2022lora}
Edward~J. Hu, Yelong Shen, Phillip Wallis, Zeyuan Allen-Zhu, Yuanzhi Li, Shean
  Wang, Lu Wang, and Weizhu Chen.
\newblock {LoRA}: Low-rank adaptation of large language models.
\newblock In \emph{International Conference on Learning Representations
  (ICLR)}, 2022.

\bibitem[{Hugging Face Diffusers team}(2023)]{diffusers2023sdxlinpaint}
{Hugging Face Diffusers team}.
\newblock {Stable Diffusion XL} inpainting 0.1.
\newblock Model released on the Hugging Face Hub
  (diffusers/stable-diffusion-xl-1.0-inpainting-0.1), 2023.

\bibitem[Ilharco et~al.(2023)Ilharco, Ribeiro, Wortsman, Gururangan, Schmidt,
  Hajishirzi, and Farhadi]{ilharco2023task}
Gabriel Ilharco, Marco~T{\'u}lio Ribeiro, Mitchell Wortsman, Suchin Gururangan,
  Ludwig Schmidt, Hannaneh Hajishirzi, and Ali Farhadi.
\newblock Editing models with task arithmetic.
\newblock In \emph{International Conference on Learning Representations
  (ICLR)}, 2023.

\bibitem[Ilyas et~al.(2022)Ilyas, Park, Engstrom, Leclerc, and
  Madry]{ilyas2022datamodels}
Andrew Ilyas, Sung~Min Park, Logan Engstrom, Guillaume Leclerc, and Aleksander
  Madry.
\newblock {Datamodels}: Understanding predictions with data and data with
  predictions.
\newblock In \emph{International Conference on Machine Learning (ICML)}, 2022.

\bibitem[Koh and Liang(2017)]{koh2017influence}
Pang~Wei Koh and Percy Liang.
\newblock Understanding black-box predictions via influence functions.
\newblock In \emph{International Conference on Machine Learning (ICML)}, 2017.

\bibitem[Kumari et~al.(2023)Kumari, Zhang, Wang, Shechtman, Zhang, and
  Zhu]{kumari2023ablating}
Nupur Kumari, Bingliang Zhang, Sheng-Yu Wang, Eli Shechtman, Richard Zhang, and
  Jun-Yan Zhu.
\newblock Ablating concepts in text-to-image diffusion models.
\newblock In \emph{IEEE/CVF International Conference on Computer Vision
  (ICCV)}, 2023.

\bibitem[Kuo et~al.(2025)Kuo, Setlur, Srinivas, Raghunathan, and
  Smith]{kuo2025siftmasks}
Kevin Kuo, Amrith Setlur, Kartik Srinivas, Aditi Raghunathan, and Virginia
  Smith.
\newblock Exact unlearning of finetuning data via model merging at scale.
\newblock \emph{arXiv preprint arXiv:2504.04626}, 2025.

\bibitem[Kwon et~al.(2024)Kwon, Wu, Wu, and Zou]{kwon2024datainf}
Yongchan Kwon, Eric Wu, Kevin Wu, and James Zou.
\newblock {DataInf}: Efficiently estimating data influence in {LoRA}-tuned
  {LLMs} and diffusion models.
\newblock In \emph{International Conference on Learning Representations
  (ICLR)}, 2024.

\bibitem[Lee et~al.(2024)Lee, Kwak, Sohn, and Shin]{lee2024dco}
Kyungmin Lee, Sangkyung Kwak, Kihyuk Sohn, and Jinwoo Shin.
\newblock Direct consistency optimization for robust customization of
  text-to-image diffusion models.
\newblock In \emph{Advances in Neural Information Processing Systems
  (NeurIPS)}, 2024.

\bibitem[Li et~al.(2022)Li, Gururangan, Dettmers, Lewis, Althoff, Smith, and
  Zettlemoyer]{li2022btm}
Margaret Li, Suchin Gururangan, Tim Dettmers, Mike Lewis, Tim Althoff, Noah~A.
  Smith, and Luke Zettlemoyer.
\newblock {Branch-Train-Merge}: Embarrassingly parallel training of expert
  language models.
\newblock \emph{arXiv preprint arXiv:2208.03306}, 2022.

\bibitem[Lin et~al.(2025{\natexlab{a}})Lin, Lu, Kim, and Lee]{lin2025crediting}
Chris Lin, Mingyu Lu, Chanwoo Kim, and Su-In Lee.
\newblock An efficient framework for crediting data contributors of diffusion
  models.
\newblock In \emph{International Conference on Learning Representations
  (ICLR)}, 2025{\natexlab{a}}.

\bibitem[Lin et~al.(2025{\natexlab{b}})Lin, Tao, Dong, and Xu]{lin2025das}
Jinxu Lin, Linwei Tao, Minjing Dong, and Chang Xu.
\newblock Diffusion attribution score: Evaluating training data influence in
  diffusion models.
\newblock In \emph{International Conference on Learning Representations
  (ICLR)}, 2025{\natexlab{b}}.

\bibitem[Lin et~al.(2014)Lin, Maire, Belongie, Hays, Perona, Ramanan,
  Doll{\'a}r, and Zitnick]{lin2014coco}
Tsung-Yi Lin, Michael Maire, Serge Belongie, James Hays, Pietro Perona, Deva
  Ramanan, Piotr Doll{\'a}r, and C.~Lawrence Zitnick.
\newblock Microsoft {COCO}: Common objects in context.
\newblock In \emph{European Conference on Computer Vision (ECCV)}, 2014.

\bibitem[Liu et~al.(2022)Liu, Li, Du, Torralba, and
  Tenenbaum]{liu2022composable}
Nan Liu, Shuang Li, Yilun Du, Antonio Torralba, and Joshua~B. Tenenbaum.
\newblock Compositional visual generation with composable diffusion models.
\newblock In \emph{European Conference on Computer Vision (ECCV)}, 2022.

\bibitem[Luo et~al.(2024)Luo, Wong, Trabucco, Huang, Gonzalez, Chen,
  Salakhutdinov, and Stoica]{luo2024stylus}
Michael Luo, Justin Wong, Brandon Trabucco, Yanping Huang, Joseph~E. Gonzalez,
  Zhifeng Chen, Ruslan Salakhutdinov, and Ion Stoica.
\newblock Stylus: Automatic adapter selection for diffusion models.
\newblock In \emph{Advances in Neural Information Processing Systems
  (NeurIPS)}, 2024.

\bibitem[Min et~al.(2024)Min, Gururangan, Wallace, Shi, Hajishirzi, Smith, and
  Zettlemoyer]{min2024silo}
Sewon Min, Suchin Gururangan, Eric Wallace, Weijia Shi, Hannaneh Hajishirzi,
  Noah~A. Smith, and Luke Zettlemoyer.
\newblock {SILO} language models: Isolating legal risk in a nonparametric
  datastore.
\newblock In \emph{International Conference on Learning Representations
  (ICLR)}, 2024.

\bibitem[Mlodozeniec et~al.(2025)Mlodozeniec, Eschenhagen, Bae, Immer, Krueger,
  and Turner]{mlodozeniec2025influence}
Bruno Mlodozeniec, Runa Eschenhagen, Juhan Bae, Alexander Immer, David Krueger,
  and Richard Turner.
\newblock Influence functions for scalable data attribution in diffusion
  models.
\newblock In \emph{International Conference on Learning Representations
  (ICLR)}, 2025.

\bibitem[Morreale et~al.(2025)Morreale, Hutiri, Serr{\`a}, Xiang, and
  Mitsufuji]{morreale2025attribution}
Fabio Morreale, Wiebke Hutiri, Joan Serr{\`a}, Alice Xiang, and Yuki Mitsufuji.
\newblock Attribution-by-design: Ensuring inference-time provenance in
  generative music systems.
\newblock \emph{arXiv preprint arXiv:2510.08062}, 2025.

\bibitem[Muqeeth et~al.(2024)Muqeeth, Liu, Liu, and
  Raffel]{muqeeth2024phatgoose}
Mohammed Muqeeth, Haokun Liu, Yufan Liu, and Colin Raffel.
\newblock Learning to route among specialized experts for zero-shot
  generalization.
\newblock In \emph{International Conference on Machine Learning (ICML)}, 2024.

\bibitem[Murata et~al.(2026)Murata, Takida, Lai, Uesaka, Nguyen, Ermon, and
  Mitsufuji]{guda2026}
Naoki Murata, Yuhta Takida, Chieh-Hsin Lai, Toshimitsu Uesaka, Bac Nguyen,
  Stefano Ermon, and Yuki Mitsufuji.
\newblock {GUDA}: Counterfactual group-wise training data attribution for
  diffusion models via unlearning.
\newblock In \emph{International Conference on Machine Learning (ICML)}, 2026.

\bibitem[Oquab et~al.(2024)Oquab, Darcet, Moutakanni, Vo, Szafraniec, Khalidov,
  Fernandez, Haziza, Massa, El-Nouby, et~al.]{oquab2024dinov2}
Maxime Oquab, Timoth{\'e}e Darcet, Th{\'e}o Moutakanni, Huy Vo, Marc
  Szafraniec, Vasil Khalidov, Pierre Fernandez, Daniel Haziza, Francisco Massa,
  Alaaeldin El-Nouby, et~al.
\newblock {DINOv2}: Learning robust visual features without supervision.
\newblock \emph{Transactions on Machine Learning Research (TMLR)}, 2024.

\bibitem[Ostapenko et~al.(2024)Ostapenko, Su, Ponti, Charlin, Le~Roux, Caccia,
  and Sordoni]{ostapenko2024arrow}
Oleksiy Ostapenko, Zhan Su, Edoardo Ponti, Laurent Charlin, Nicolas Le~Roux,
  Lucas Caccia, and Alessandro Sordoni.
\newblock Towards modular {LLMs} by building and reusing a library of {LoRAs}.
\newblock In \emph{International Conference on Machine Learning (ICML)}, 2024.

\bibitem[Ouyang et~al.(2025)Ouyang, Li, and Hou]{ouyang2025klora}
Ziheng Ouyang, Zhen Li, and Qibin Hou.
\newblock {K-LoRA}: Unlocking training-free fusion of any subject and style
  {LoRAs}.
\newblock In \emph{IEEE/CVF Conference on Computer Vision and Pattern
  Recognition (CVPR)}, 2025.

\bibitem[Park et~al.(2023)Park, Georgiev, Ilyas, Leclerc, and
  Madry]{park2023trak}
Sung~Min Park, Kristian Georgiev, Andrew Ilyas, Guillaume Leclerc, and
  Aleksander Madry.
\newblock {TRAK}: Attributing model behavior at scale.
\newblock In \emph{International Conference on Machine Learning (ICML)}, 2023.

\bibitem[Peng et~al.(2026)Peng, Zheng, Yang, Huang, Yan, Liu, and
  Chen]{tara2026}
Yuqi Peng, Lingtao Zheng, Yufeng Yang, Yi Huang, Mingfu Yan, Jianzhuang Liu,
  and Shifeng Chen.
\newblock {TARA}: Token-aware {LoRA} for composable personalization in
  diffusion models.
\newblock In \emph{AAAI Conference on Artificial Intelligence (AAAI)}, 2026.

\bibitem[Po et~al.(2024)Po, Yang, Aberman, and Wetzstein]{po2024orthogonal}
Ryan Po, Guandao Yang, Kfir Aberman, and Gordon Wetzstein.
\newblock Orthogonal adaptation for modular customization of diffusion models.
\newblock In \emph{IEEE/CVF Conference on Computer Vision and Pattern
  Recognition (CVPR)}, 2024.

\bibitem[Podell et~al.(2024)Podell, English, Lacey, Blattmann, Dockhorn,
  M{\"u}ller, Penna, and Rombach]{podell2024sdxl}
Dustin Podell, Zion English, Kyle Lacey, Andreas Blattmann, Tim Dockhorn, Jonas
  M{\"u}ller, Joe Penna, and Robin Rombach.
\newblock {SDXL}: Improving latent diffusion models for high-resolution image
  synthesis.
\newblock In \emph{International Conference on Learning Representations
  (ICLR)}, 2024.

\bibitem[Radford et~al.(2021)Radford, Kim, Hallacy, Ramesh, Goh, Agarwal,
  Sastry, Askell, Mishkin, Clark, Krueger, and Sutskever]{radford2021clip}
Alec Radford, Jong~Wook Kim, Chris Hallacy, Aditya Ramesh, Gabriel Goh,
  Sandhini Agarwal, Girish Sastry, Amanda Askell, Pamela Mishkin, Jack Clark,
  Gretchen Krueger, and Ilya Sutskever.
\newblock Learning transferable visual models from natural language
  supervision.
\newblock In \emph{International Conference on Machine Learning (ICML)}, 2021.

\bibitem[Rombach et~al.(2022)Rombach, Blattmann, Lorenz, Esser, and
  Ommer]{rombach2022ldm}
Robin Rombach, Andreas Blattmann, Dominik Lorenz, Patrick Esser, and Bj{\"o}rn
  Ommer.
\newblock High-resolution image synthesis with latent diffusion models.
\newblock In \emph{IEEE/CVF Conference on Computer Vision and Pattern
  Recognition (CVPR)}, 2022.

\bibitem[Ruiz et~al.(2023)Ruiz, Li, Jampani, Pritch, Rubinstein, and
  Aberman]{ruiz2023dreambooth}
Nataniel Ruiz, Yuanzhen Li, Varun Jampani, Yael Pritch, Michael Rubinstein, and
  Kfir Aberman.
\newblock {DreamBooth}: Fine tuning text-to-image diffusion models for
  subject-driven generation.
\newblock In \emph{IEEE/CVF Conference on Computer Vision and Pattern
  Recognition (CVPR)}, 2023.

\bibitem[Schuhmann(2022)]{schuhmann2022aesthetic}
Christoph Schuhmann.
\newblock Improved aesthetic predictor.
\newblock GitHub repository christophschuhmann/improved-aesthetic-predictor,
  2022.

\bibitem[Schuhmann et~al.(2022)Schuhmann, Beaumont, Vencu, Gordon, Wightman,
  Cherti, Coombes, Katta, Mullis, Wortsman, Schramowski, Kundurthy, Crowson,
  Schmidt, Kaczmarczyk, and Jitsev]{schuhmann2022laion5b}
Christoph Schuhmann, Romain Beaumont, Richard Vencu, Cade Gordon, Ross
  Wightman, Mehdi Cherti, Theo Coombes, Aarush Katta, Clayton Mullis, Mitchell
  Wortsman, Patrick Schramowski, Srivatsa Kundurthy, Katherine Crowson, Ludwig
  Schmidt, Robert Kaczmarczyk, and Jenia Jitsev.
\newblock {LAION-5B}: An open large-scale dataset for training next generation
  image-text models.
\newblock In \emph{Advances in Neural Information Processing Systems (NeurIPS),
  Datasets and Benchmarks Track}, 2022.

\bibitem[Serr{\`a} et~al.(2026)Serr{\`a}, Goswami, Morreale, Liao, and
  Mitsufuji]{serra2026mucs}
Joan Serr{\`a}, Dipam Goswami, Fabio Morreale, Wei-Hsiang Liao, and Yuki
  Mitsufuji.
\newblock Training data attribution in diffusion models via mirrored unlearning
  and noise-consistent skew.
\newblock In \emph{Advances in Neural Information Processing Systems
  (NeurIPS)}, 2026.

\bibitem[Shah et~al.(2024)Shah, Ruiz, Cole, Lu, Lazebnik, Li, and
  Jampani]{shah2024ziplora}
Viraj Shah, Nataniel Ruiz, Forrester Cole, Erika Lu, Svetlana Lazebnik,
  Yuanzhen Li, and Varun Jampani.
\newblock {ZipLoRA}: Any subject in any style by effectively merging {LoRAs}.
\newblock In \emph{European Conference on Computer Vision (ECCV)}, 2024.

\bibitem[Shapley(1953)]{shapley1953value}
Lloyd~S. Shapley.
\newblock A value for $n$-person games.
\newblock In \emph{Contributions to the Theory of Games, Volume II}, pages
  307--317. Princeton University Press, 1953.

\bibitem[Shi et~al.(2025)Shi, Bhagia, Farhat, Muennighoff, Morrison, Walsh,
  Schwenk, Longpre, Poznanski, Ettinger, et~al.]{shi2025flexolmo}
Weijia Shi, Akshita Bhagia, Kevin Farhat, Niklas Muennighoff, Jacob Morrison,
  Evan~Pete Walsh, Dustin Schwenk, Shayne Longpre, Jake Poznanski, Allyson
  Ettinger, et~al.
\newblock {FlexOLMo}: Open language models for flexible data use.
\newblock In \emph{Advances in Neural Information Processing Systems
  (NeurIPS)}, 2025.

\bibitem[Skreta et~al.(2025)Skreta, Atanackovic, Bose, Tong, and
  Neklyudov]{skreta2025superdiff}
Marta Skreta, Lazar Atanackovic, Avishek~Joey Bose, Alexander Tong, and Kirill
  Neklyudov.
\newblock The superposition of diffusion models using the {It{\^o}} density
  estimator.
\newblock In \emph{International Conference on Learning Representations
  (ICLR)}, 2025.

\bibitem[Somepalli et~al.(2023)Somepalli, Singla, Goldblum, Geiping, and
  Goldstein]{somepalli2023forgery}
Gowthami Somepalli, Vasu Singla, Micah Goldblum, Jonas Geiping, and Tom
  Goldstein.
\newblock Diffusion art or digital forgery? {Investigating} data replication in
  diffusion models.
\newblock In \emph{IEEE/CVF Conference on Computer Vision and Pattern
  Recognition (CVPR)}, 2023.

\bibitem[Somepalli et~al.(2024)Somepalli, Gupta, Gupta, Palta, Goldblum,
  Geiping, Shrivastava, and Goldstein]{somepalli2024csd}
Gowthami Somepalli, Anubhav Gupta, Kamal Gupta, Shramay Palta, Micah Goldblum,
  Jonas Geiping, Abhinav Shrivastava, and Tom Goldstein.
\newblock Investigating style similarity in diffusion models.
\newblock In \emph{European Conference on Computer Vision (ECCV)}, 2024.

\bibitem[Tsoumplekas et~al.(2026)Tsoumplekas, Bounareli, and
  Argyriou]{tsoumplekas2026prompt}
Georgios Tsoumplekas, Stella Bounareli, and Vasileios Argyriou.
\newblock Training-free multi-concept {LoRA} composition with prompt-aware
  weighting.
\newblock In \emph{IEEE International Conference on Automatic Face and Gesture
  Recognition (FG)}, 2026.

\bibitem[Vyas et~al.(2023)Vyas, Kakade, and Barak]{vyas2023provable}
Nikhil Vyas, Sham Kakade, and Boaz Barak.
\newblock On provable copyright protection for generative models.
\newblock In \emph{International Conference on Machine Learning (ICML)}, 2023.

\bibitem[Wang et~al.(2024{\natexlab{a}})Wang, Deng, Chiba-Okabe, Barak, and
  Su]{wang2024economic}
Jiachen~T. Wang, Zhun Deng, Hiroaki Chiba-Okabe, Boaz Barak, and Weijie~J. Su.
\newblock An economic solution to copyright challenges of generative {AI}.
\newblock \emph{arXiv preprint arXiv:2404.13964}, 2024{\natexlab{a}}.

\bibitem[Wang et~al.(2023)Wang, Efros, Zhu, and Zhang]{wang2023evaluating}
Sheng-Yu Wang, Alexei~A. Efros, Jun-Yan Zhu, and Richard Zhang.
\newblock Evaluating data attribution for text-to-image models.
\newblock In \emph{IEEE/CVF International Conference on Computer Vision
  (ICCV)}, 2023.

\bibitem[Wang et~al.(2024{\natexlab{b}})Wang, Hertzmann, Efros, Zhu, and
  Zhang]{wang2024unlearning}
Sheng-Yu Wang, Aaron Hertzmann, Alexei~A. Efros, Jun-Yan Zhu, and Richard
  Zhang.
\newblock Data attribution for text-to-image models by unlearning synthesized
  images.
\newblock In \emph{Advances in Neural Information Processing Systems
  (NeurIPS)}, 2024{\natexlab{b}}.

\bibitem[Wang et~al.(2025)Wang, Hertzmann, Efros, Zhang, and
  Zhu]{wang2025fastgda}
Sheng-Yu Wang, Aaron Hertzmann, Alexei~A. Efros, Richard Zhang, and Jun-Yan
  Zhu.
\newblock Fast data attribution for text-to-image models.
\newblock In \emph{Advances in Neural Information Processing Systems
  (NeurIPS)}, 2025.

\bibitem[Wortsman et~al.(2022)Wortsman, Ilharco, Gadre, Roelofs, Gontijo-Lopes,
  Morcos, Namkoong, Farhadi, Carmon, Kornblith, and Schmidt]{wortsman2022soups}
Mitchell Wortsman, Gabriel Ilharco, Samir~Yitzhak Gadre, Rebecca Roelofs,
  Raphael Gontijo-Lopes, Ari~S. Morcos, Hongseok Namkoong, Ali Farhadi, Yair
  Carmon, Simon Kornblith, and Ludwig Schmidt.
\newblock Model soups: averaging weights of multiple fine-tuned models improves
  accuracy without increasing inference time.
\newblock In \emph{International Conference on Machine Learning (ICML)}, 2022.

\bibitem[Wu et~al.(2025)Wu, Le, Hayat, and Harandi]{wu2025erasediff}
Jing Wu, Trung Le, Munawar Hayat, and Mehrtash Harandi.
\newblock Erasing undesirable influence in diffusion models.
\newblock In \emph{IEEE/CVF Conference on Computer Vision and Pattern
  Recognition (CVPR)}, 2025.

\bibitem[Wu et~al.(2024)Wu, Huang, and Wei]{wu2024mole}
Xun Wu, Shaohan Huang, and Furu Wei.
\newblock Mixture of {LoRA} experts.
\newblock In \emph{International Conference on Learning Representations
  (ICLR)}, 2024.

\bibitem[Yadav et~al.(2023)Yadav, Tam, Choshen, Raffel, and
  Bansal]{yadav2023ties}
Prateek Yadav, Derek Tam, Leshem Choshen, Colin Raffel, and Mohit Bansal.
\newblock {TIES-Merging}: Resolving interference when merging models.
\newblock In \emph{Advances in Neural Information Processing Systems
  (NeurIPS)}, 2023.

\bibitem[Yang et~al.(2025)Yang, Wang, Peng, Song, Chen, Li, Yang, Lu, Cai, He,
  Wu, and Liu]{yang2025loracomposer}
Yang Yang, Wen Wang, Liang Peng, Chaotian Song, Yao Chen, Hengjia Li, Xiaolong
  Yang, Qinglin Lu, Deng Cai, Xiaofei He, Boxi Wu, and Wei Liu.
\newblock {LoRA-Composer}: Leveraging low-rank adaptation for multi-concept
  customization in training-free diffusion models.
\newblock \emph{IEEE Transactions on Image Processing (TIP)}, 34:\penalty0
  8145--8158, 2025.

\bibitem[Zhao et~al.(2024)Zhao, Gan, Wang, Zhou, Yang, Kuang, and
  Wu]{zhao2024loraretriever}
Ziyu Zhao, Leilei Gan, Guoyin Wang, Wangchunshu Zhou, Hongxia Yang, Kun Kuang,
  and Fei Wu.
\newblock {LoraRetriever}: Input-aware {LoRA} retrieval and composition for
  mixed tasks in the wild.
\newblock In \emph{Findings of the Association for Computational Linguistics
  (ACL)}, pages 4447--4462, 2024.

\bibitem[Zheng et~al.(2026)Zheng, Tai, and Yeh]{zheng2026designing}
Amber~Yijia Zheng, Yu-Shan Tai, and Raymond~A. Yeh.
\newblock Designing to forget: Deep semi-parametric models for unlearning.
\newblock In \emph{IEEE/CVF Conference on Computer Vision and Pattern
  Recognition (CVPR)}, 2026.

\bibitem[Zheng et~al.(2024)Zheng, Pang, Du, Jiang, and Lin]{zheng2024dtrak}
Xiaosen Zheng, Tianyu Pang, Chao Du, Jing Jiang, and Min Lin.
\newblock Intriguing properties of data attribution on diffusion models.
\newblock In \emph{International Conference on Learning Representations
  (ICLR)}, 2024.

\bibitem[Zhong et~al.(2024)Zhong, Shen, Wang, Lu, Jiao, Ouyang, Yu, Han, and
  Chen]{zhong2024multilora}
Ming Zhong, Yelong Shen, Shuohang Wang, Yadong Lu, Yizhu Jiao, Siru Ouyang,
  Donghan Yu, Jiawei Han, and Weizhu Chen.
\newblock {Multi-LoRA} composition for image generation.
\newblock \emph{Transactions on Machine Learning Research (TMLR)}, 2024.

\end{thebibliography}
